\documentclass{article}

\usepackage{arxiv}
\usepackage[utf8]{inputenc} 
\usepackage[T1]{fontenc}    
\usepackage{hyperref}       
\usepackage{url}            
\usepackage{booktabs}       
\usepackage{amsfonts}       
\usepackage{nicefrac}       
\usepackage{microtype}      
\usepackage{lipsum}		
\usepackage{graphicx}
\usepackage{natbib}
\usepackage{doi}

\usepackage{amsmath}
\usepackage{algorithm}
\usepackage{amssymb}
\usepackage{tabularx}
\usepackage{ragged2e} 

\title{Autonomous Navigation at the Nano-Scale: Algorithms, Architectures, and Constraints}

\author{ 
    Mahmud S. Zango \\ 
    James Watt School of Engineering\\
    University of Glasgow\\
    Glasgow G12 8QQ, UK \\
    \texttt{m.zango.1@research.gla.ac.uk} \\
    \And
    \href{https://orcid.org/0000-0001-9057-5649}{\includegraphics[scale=0.06]{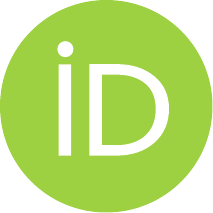}\hspace{1mm}Jianglin Lan}\thanks{Correspondence: \texttt{Jianglin.Lan@glasgow.ac.uk}} \\
    James Watt School of Engineering\\
    University of Glasgow\\
    Glasgow G12 8QQ, UK \\
    \texttt{Jianglin.Lan@glasgow.ac.uk} \\
}
\date{}
\renewcommand{\shorttitle}{Autonomous Navigation at Nano-Scale} 

\hypersetup{
pdftitle={Autonomous navigation at the nano-scale: algorithms, architectures, and constraints},
pdfsubject={Robotics, Edge AI, Control Theory},
pdfauthor={Mahmud S. Zango, Jianglin Lan},
pdfkeywords={Nano-UAVs, Edge AI, Neuromorphic Control, SWaP Constraints, Sim-to-Real Transfer, TinyML},
}
\begin{document}
\maketitle

\begin{abstract}
	Autonomous navigation for nano-scale unmanned aerial vehicles (nano-UAVs) is governed by extreme Size, Weight, and Power (SWaP) constraints (with the weight < 50 g and sub-100 mW onboard processor), distinguishing it fundamentally from standard robotic paradigms. 
    This review synthesizes the state-of-the-art in sensing, computing, and control architectures designed specifically for these sub-100mW computational envelopes. We critically analyse the transition from classical geometry-based methods to emerging "Edge AI" paradigms, including quantized deep neural networks deployed on ultra-low-power System-on-Chips (SoCs) and neuromorphic event-based control. Beyond algorithms, we evaluate the hardware-software co-design requisite for autonomy, covering advancements in dense optical flow, optimized Simultaneous Localization and Mapping (SLAM), and learning-based flight control. While significant progress has been observed in visual navigation and relative pose estimation, our analysis reveals persistent gaps in long-term endurance, robust obstacle avoidance in dynamic environments, and the "Sim-to-Real" transfer of reinforcement learning policies. This survey provides a roadmap for bridging these gaps, advocating for hybrid architectures that fuse lightweight classical control with data-driven perception to enable fully autonomous, agile nano-UAVs in GPS-denied environments.
\end{abstract}

\keywords{Nano-UAVs \and Edge AI \and Neuromorphic control \and SWaP constraints \and Sim-to-Real transfer \and TinyML}

\section{Introduction}

\begin{figure*}[t!]
    \centering
    \includegraphics[width=\textwidth]{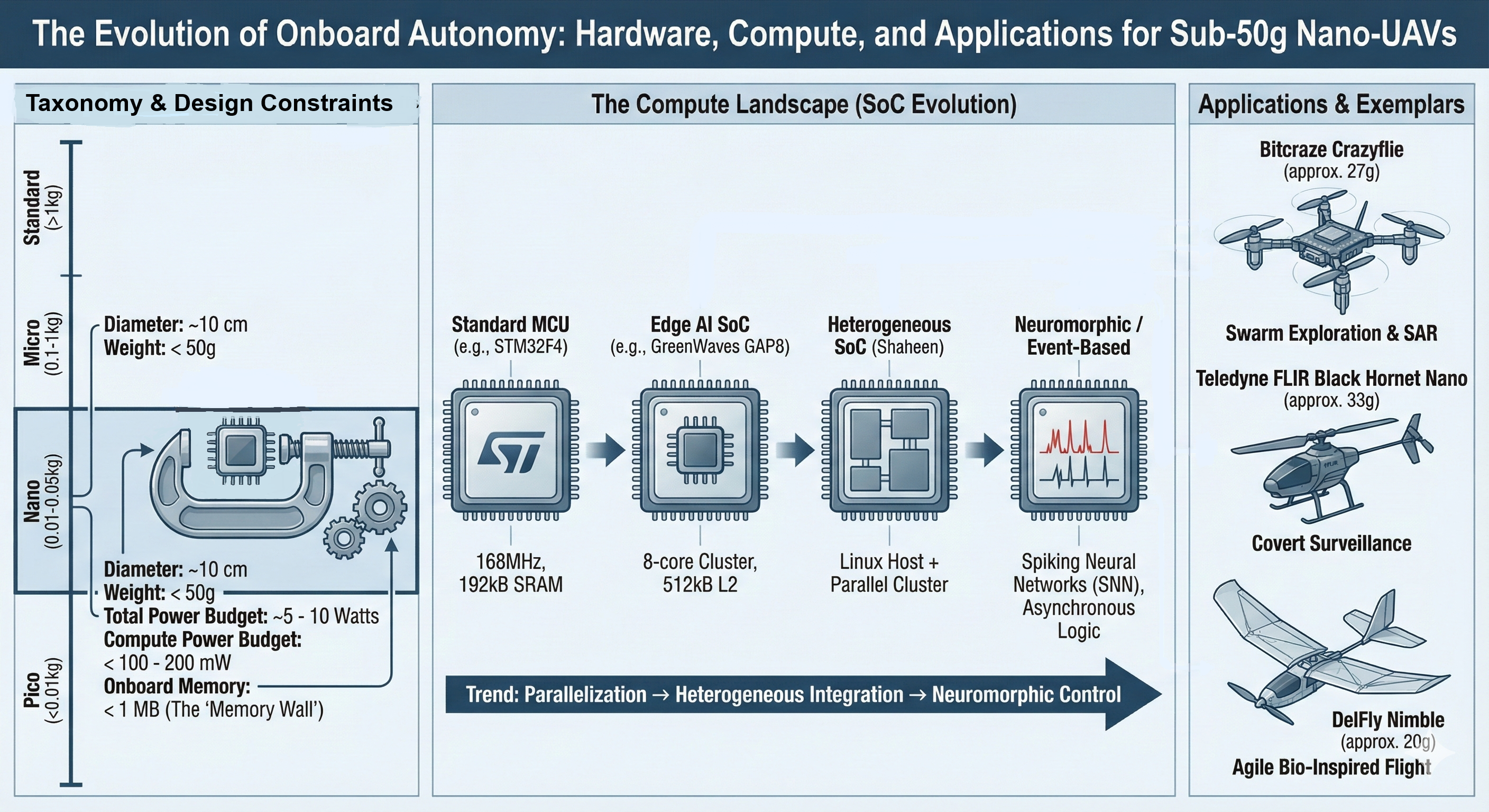} 
    \caption{\textbf{The evolution of onboard autonomy for nano-UAVs.} Left: Severe Size, Weight, and Power (SWaP) constraints that limit sub-50g platforms. Center: The progression of system-on-chip (SoC) architectures from simple microcontrollers to heterogeneous and neuromorphic processors to address the computational gap. Right: Representative platforms and application domains enabled by these advances.}
    \label{fig:overview}
\end{figure*}

The aerospace domain is witnessing a fundamental paradigm shift, transitioning from a reliance on robust, singular large-scale platforms toward the deployment of distributed swarms of miniaturized agents. Driven by the aggressive miniaturization of electromechanical components and the rise of edge computing, this transition has enabled the proliferation of sub-50g autonomous platforms \cite{cai2014survey}. These ``Nano-Fliers'' possess the revolutionary potential to penetrate environments previously inaccessible to standard air vehicles, transforming applications ranging from search-and-rescue in collapsed infrastructure to distributed environmental sensing \cite{cai2014survey, scarciglia2025map}. However, realizing true autonomy at this scale requires overcoming severe Size, Weight, and Power (SWaP) constraints, where the total power budget is often limited to a few Watts, with only a fraction available for onboard computation \cite{palossi201964, niculescu2022energy} (see Fig. \ref{fig:overview}).

In this review, we categorize these nano-scale robots not merely by their operational domain, but by the aerodynamic regimes they exploit to achieve agility and stability. The first category, \textbf{Rotary-Wing Nano-UAVs}, is exemplified by platforms such as the Bitcraze Crazyflie ($\approx 27$g); these vehicles utilize high-speed rotors for holonomic agility and have become the standard research platform for swarm coordination and navigation in cluttered indoor environments \cite{cai2014survey, sartori2025ai, 7989376}. The second category, \textbf{Bio-Inspired Flappers}, consists of insect-scale ornithopters that exploit unsteady aerodynamics to generate lift and control. Prominent examples include the DelFly Nimble (28g) and the Harvard RoboBee (80mg), the latter mimicking the mechanics of insect flight at the pico-scale \cite{cai2014survey}. The unifying advantage of these diverse architectures is their ability to operate within highly constrained three-dimensional volumes where standard UAVs are physically excluded or operationally unsafe.

The operational reality of nano-fliers is governed by a ``Physics Gap'' where the scaling laws of aerodynamics diverge significantly from standard aviation models. Operating at low Reynolds numbers ($Re < 10^4$), air viscosity dominates inertial forces, rendering the fluid dynamic environment akin to syrup rather than the inviscid flow assumed for larger aircraft \cite{forster2015system, helps2022liquid}. This necessitates reliance on unsteady aerodynamics for lift and stability, making vehicles highly susceptible to unmodeled disturbances such as wind gusts and motor-induced vibrations \cite{bernini2021few, bauersfeld2021neurobem}. These physical disparities exacerbate the ``Sim-to-Real'' gap, as standard simulation environments often fail to capture the stochastic nonlinearities of nano-scale physics, leading to control policies that perform robustly \textit{in silico} but fail in reality \cite{huang2023dattdeepadaptivetrajectory}.

Simultaneously, the ``SWaP Gap'' imposes severe design constraints that prevent the deployment of conventional autonomy stacks. The high energetic cost of flight at small scales, coupled with battery capacities often below 250 mAh, limits operational endurance to minutes for rotary-wing vehicles and mere seconds for flapping-wing prototypes \cite{kooi2021inclined, giernacki2017crazyflie, e2fbb835-8278-3148-8a69-8aad65c23c4e}. This scarcity restricts the onboard computational power budget to approximately 5--15\% of the total power (often $< 100$ mW), forcing reliance on resource-constrained Microcontroller Units (MCUs) or ultra-low-power SoCs rather than the GPUs utilized on standard-class UAVs \cite{valente2024heterogeneous, palossi201964, muller2023fully}. Consequently, computationally expensive algorithms such as high-fidelity Visual Simultaneous Localization and Mapping (VSLAM) are infeasible onboard \cite{valente2024heterogeneous, muller2023fully}. Furthermore, strict payload limitations ($< 15$ g) exclude high-precision sensors like LiDAR, necessitating navigation solutions that rely on noisy, low-fidelity data from sparse Time-of-Flight (ToF) sensors or optical flow \cite{scarciglia2025map, friess2024fully,neuman2022tiny}.

To circumvent these physical and computational barriers, the field is converging toward ``lightweight'' solutions that prioritize efficiency and biomimicry over raw processing power. Foremost among these is \textbf{Neuromorphic Sensing}, particularly the use of event-based cameras. Unlike standard frame-based cameras that suffer from motion blur and data redundancy, event cameras offer microsecond-level latency and high dynamic range at low power, enabling the high-speed manoeuvres essential for agile nano-fliers \cite{bian2023colibriuav, kuhne2023fast}. Complementing this is the adoption of \textbf{Bio-Inspired Control}, which mimics the optical flow processing of insect compound eyes to facilitate navigation without the weight penalty of stereo-vision systems \cite{paredes2024fully, 10.1007/978-3-031-63596-0_37}. Furthermore, to overcome individual limitations in payload and range, \textbf{Swarm Intelligence} leverages decentralized algorithms allowing platforms to achieve complex mission goals through local interaction rules rather than computationally expensive centralized planning \cite{7989376, uppaluru2023multi, chen2024learning}. This review analyzes the ``Onboard Autonomy Stack'' required to operationalize these paradigms. We emphasize the necessity of holistic co-design, where airframe limitations dictate algorithm design, and conversely, autonomy requirements drive hardware specialization \cite{e2fbb835-8278-3148-8a69-8aad65c23c4e,kuhne2023fast}.

Distinct from prior literature, this review addresses the specific intersection of extreme SWaP constraints and onboard algorithmic autonomy. While Cai et al. \cite{cai2014survey} provided a foundational survey of small-scale platforms, their work primarily catalogs aerodynamic configurations and classical control strategies relevant to the pre-edge-AI era. Similarly, while Neuman et al. \cite{neuman2022tiny} effectively highlight the challenges of ``Tiny Robot Learning,'' their scope encompasses general micro-robotics (including terrestrial quadrupeds) with a primary emphasis on machine learning inference paradigms. In contrast, this review specifically targets the \textit{Nano-UAV} domain, bridging the gap between hardware limitations and the full autonomy stack. We exclusively focus on the computational and algorithmic architectures required to close the perception-actuation loop entirely onboard under sub-100\,mW constraints, moving beyond isolated subsystems to holistic autonomy.

\subsection{Review Methodology and Scope}
To provide a comprehensive analysis of the state-of-the-art, we surveyed literature from IEEE Xplore, ScienceDirect, and arXiv repositories, covering the period from 2014--2025. Special emphasis was placed on the post-2019 era, which marks the emergence of heterogeneous edge-AI accelerators (e.g., RISC-V GAP8) in the nano-robotics domain. The selection process prioritized experimental validity under strict Size, Weight, and Power (SWaP) constraints. Publications were selected based on three rigorous inclusion criteria:

\begin{itemize}
    \item \textbf{Form Factor:} We restricted our scope to ``nano'' class aerial platforms, defined here as vehicles with a total weight of $<50$\,g and a diameter of $<10$\,cm (e.g., Bitcraze Crazyflie, perceptually-enabled insect-scale robots). General Micro Aerial Vehicle (MAV) literature ($>100$\,g) was excluded unless it introduced algorithmic primitives explicitly adapted for extreme quantization or low-inertia dynamics.
    
    \item \textbf{Onboard/Off-Board Autonomy:} Priority was given to methodologies demonstrating fully onboard processing, where perception, state estimation, and control run locally on the microcontroller or/and edge accelerator. Approaches relying on external motion capture systems (e.g., Vicon/Optitrack) or offboard computation via radio streaming were excluded, except to serve as ground-truth baselines, due to their non-viability in real-world deployment.
    
    \item \textbf{Algorithmic Efficiency:} The review targets works contributing to ``TinyML,'' spiking neural networks (SNNs), and lightweight control strategies (e.g., quantized DNNs, depth-based obstacle avoidance). We focused on algorithms capable of closing the perception-actuation loop within the milliwatt-range power envelopes of embedded SoCs, excluding heavy visual-inertial odometry (VIO) pipelines requiring desktop-class GPUs.
\end{itemize}
 
\subsection{Organization of the Review}
The remainder of this paper is organized as follows: Section ~\ref{sec:hardware} reviews the hardware platforms and low-power sensing architectures fundamental to the nano-scale. Section ~\ref{sec:algorithms} dissects the ``Onboard Autonomy Stack'', covering efficient machine learning perception, optimization-based planning, robust flight control, and decentralized swarm coordination, alongside the software toolchains required for deployment. Section ~\ref{sec:challenges} critically analyses the coupled physical, computational, and bandwidth constraints that define the system’s operational envelope.  Section ~\ref{sec:applications} evaluates the feasibility of these systems across diverse application domains, from industrial inspection to precision agriculture. Finally, Section ~\ref{sec:open challenges} identifies persistent research gaps, and Section 7 concludes with a roadmap for future developments, advocating for a paradigm shift toward bio-inspired and neuromorphic architectures.

\section{Hardware Platforms for Sub-50g Nano-UAVs}
\label{sec:hardware}

The computational architecture of nano-UAVs is strictly governed by SWaP constraints, which fundamentally differentiate these platforms from standard robotic systems. Typically, the total power budget for sub-50g vehicles is limited to a few Watts, with the avionics and computational subsystem restricted to merely 5--15\% of this total, equating to approximately 100--200 mW \cite{valente2024heterogeneous, palossi201964}. This scarcity necessitates a careful selection of computing substrates that balances the demand for autonomy against the physical realities of flight endurance.

\subsection{Processing Architectures}
The baseline computing substrate for research-grade nano-quadrotors (exemplified by the widely adopted Bitcraze Crazyflie 2.1) relies on standard Microcontroller Units (MCUs) such as the STM32F405 (ARM Cortex-M4). Operating at 168 MHz with 192 kB of SRAM, these MCUs are optimized for low-latency control loops, including state estimation and PID control, as well as basic sensor interfacing \cite{giernacki2017crazyflie, palossi2021fully}. However, the sequential processing nature of standard MCUs creates a significant bottleneck for advanced autonomy. While these processors can execute basic optical flow algorithms or simplistic deep learning models (such as Z-axis estimation requiring $<3$ kMAC/frame \cite{cereda2021improving}), they lack the throughput necessary for complex perception tasks like Simultaneous Localization and Mapping (SLAM) or dense object detection, which demand billions of operations per second \cite{palossi201964}.

To overcome the ``von Neumann bottleneck'' inherent in sequential MCUs without exceeding the strict power envelope, the field has witnessed a paradigm shift toward Parallel Ultra-Low-Power (PULP) architectures, such as the GreenWaves GAP8 and GAP9 Systems-on-Chip (SoC) \cite{muller2023fully,9381618}. Unlike the single-core architecture of the STM32, the GAP8 features a heterogeneous design comprising a single ``Fabric Controller'' core for I/O management and an 8-core RISC-V ``Cluster'' acting as a programmable accelerator \cite{9381618}. This parallel architecture leverages a shared L1 scratchpad memory and Single Instruction Multiple Data (SIMD) extensions to execute Convolutional Neural Networks (CNNs) with high efficiency \cite{9381618}. For example, a GAP8-equipped AI-deck can execute the PULP-Frontnet CNN at up to 135 frames per second (fps) within an 86 mW power envelope, which is a level of performance unattainable on a standard Cortex-M4 \cite{palossi2021fully}. Beyond standard parallelism, recent research has demonstrated the efficacy of \textbf{Neuromorphic principles} on standard embedded hardware. Stroobants et al. 
\cite{stroobants2023neuromorphic}
successfully implemented an end-to-end Spiking Neural Network (SNN) for attitude control on a Cortex-M7 microcontroller (Teensy 4.0). By leveraging the sparse, binary nature of spiking networks, they demonstrated that computationally expensive floating-point multiplications can be replaced by energy-efficient integer additions, achieving 500 Hz control loops within a negligible power footprint, laying the groundwork for future fully neuromorphic autopilots.

For tasks requiring maximum computational efficiency, Application-Specific Integrated Circuits (ASICs) offer superior performance per watt compared to general-purpose programmable logic. The ``Navion'' chip \cite{suleiman2019navion} illustrates this potential; as a fully integrated Visual-Inertial Odometry (VIO) accelerator, it reduces energy consumption by orders of magnitude compared to standard CPUs. Navion is capable of processing stereo images at up to 171 fps with a peak power consumption of only 24 mW, scaling down to 2 mW at standard rates, thereby eliminating the need for off-chip storage and processing \cite{suleiman2019navion}. Further advancing this trend is the ``Shaheen''  SoC \cite{valente2024heterogeneous}, which represents the next generation of heterogeneous integration by coupling a Linux-capable 64-bit RISC-V host (RV64) with a parallel 32-bit RISC-V cluster (RV32) and HyperRAM. Unlike standard MCUs restricted to bare-metal or Real-Time Operating System (RTOS) environments, Shaheen supports full-fledged operating systems alongside real-time control within a 200 mW envelope, effectively bridging the gap between nano-drones and higher-class embedded computers \cite{valente2024heterogeneous}. Table\ref{tab:compute_landscape} provides a comprehensive comparison of these computing substrates, summarizing the trade-offs between power envelopes, memory constraints, and computational throughput across standard and nano-scale platforms.
\begin{table*}[t!]
\centering
\caption{Comparative Analysis of Compute Substrates: Standard vs. Nano-Scale (Normalized)}
\label{tab:compute_landscape}
\resizebox{\textwidth}{!}{%
\begin{tabular}{@{}llcccll@{}}
\toprule
\textbf{Device} & \textbf{Architecture} & \textbf{Power} & \textbf{Peak Throughput} & \textbf{Efficiency} & \textbf{Memory} & \textbf{Primary Role} \\
 & & \textit{(mW)} & \textit{(GOPS)} & \textit{(GOPS/W)} & & \\ \midrule
\textit{Jetson TX2} \cite{nvidia_jetson_tx2} & \textit{4$\times$A57 (GPU accel.)} & \textit{7,500} & \textit{1,300\textsuperscript{a}} & \textit{0.17} & \textit{8 GB} & \textit{Mission Comp.} \\ \midrule
STM32F4 \cite{stm32f4_web} & Cortex-M4 @ 168 MHz & 100 & $<$0.5 & $<$5 & 192 kB & Flight control \\
Teensy 4.0 \cite{10935624} & Cortex-M7 @ 600 MHz & $\sim$100 & 1.2 & $\sim$12 & 1 MB & SNN research \\
GAP8 \cite{flamand2018gap} & RISC-V Cluster (8-core) & 86 & 20.0\textsuperscript{b} & 232 & 512 kB & Edge AI \\
GAP9 \cite{10.1007/978-3-031-76424-0_52} & RISC-V Cluster (9-core) & 50 & 3.0\textsuperscript{c} & 60 & 1.5 MB & AV / AI \\
Kraken \cite{9895621} & RISC-V + SNN Accel. & 300 & 3.0 & 10 & 1 MB & Neuromorphic \\
Navion \cite{suleiman2019navion} & Custom ASIC (VIO) & 24 & Task-Specific\textsuperscript{d} & N/A & On-chip & Visual odometry \\ 
Shaheen \cite{valente2024heterogeneous} & RV64 + RV32 Cluster & 200 & 7.9\textsuperscript{c} & 39.5 & HyperRAM & Linux + RTOS \\ \bottomrule
\end{tabular}%
}
\vspace{1mm}
\raggedright
\footnotesize{
\textit{Note: Power figures represent typical operational envelopes. Efficiency is calculated as Peak Throughput / Power.} \\
\textsuperscript{a} Converted from 1.3 TFLOPs for unit consistency. \\
\textsuperscript{b} Based on 10 GMACs ($1 \text{ MAC} \approx 2 \text{ OPS}$). \\
\textsuperscript{c} Reported as GFLOPs (Floating Point). \\
\textsuperscript{d} Throughput for ASICs is application-dependent (e.g., 171 FPS for Stereo-VIO \cite{suleiman2019navion}).
}
\end{table*}
\subsection{Platform Typologies: Rotary-Wing and Bio-Inspired Systems}
In the domain of rotary-wing platforms, the Bitcraze Crazyflie 2.X serves as the \textit{de facto} open-source standard for research. Weighing 27 g and measuring 92 mm motor-to-motor, it offers approximately 7 minutes of flight time \cite{giernacki2017crazyflie, 7989376} and relies on modular expansions for advanced sensing. In contrast, the Black Hornet Nano by FLIR represents the military-grade benchmark for deployed nano-UAVs. Weighing approximately 33 g, this closed system is optimized for stealth and surveillance in varying wind conditions, distinguishing it from the modular, research-focused Crazyflie \cite{neuman2022tiny}.

Bio-inspired designs on the other hand offer an alternative approach, trading the mechanical simplicity of rotors for the agility and efficiency of flapping wings. The RoboBee, an insect-scale robot weighing approximately 80 mg, utilizes piezoelectric actuators to mimic insect flight but faces extreme power constraints that have historically required tethers \cite{e2fbb835-8278-3148-8a69-8aad65c23c4e}. Larger flapping-wing micro air vehicles like the DelFly (20g+) have demonstrated the capacity to carry stereo vision systems, proving that flapping flight can support perception payloads \cite{6907589}. Pushing the envelope of biological mimicry, the Colibri UAV \cite{bian2023colibriuav} integrates neuromorphic computing via the Kraken SoC \cite{9895621} with event-based cameras. While this platform physically sits at the upper bound of the sub-50g category (often exceeding it with battery payloads), its architecture serves as a critical benchmark for energy efficiency. By utilizing event-based processing, it achieves microsecond-level reaction times and milliwatt-scale power consumption, successfully replicating biological visual processing speeds to enable agile flight in dynamic environments \cite{bian2023colibriuav, 10935624}. 

Furthermore, novel actuation mechanisms like the Liquid-amplified Zipping Actuator (LAZA) are eliminating transmission losses entirely \cite{helps2022liquid}. Unlike traditional motors that require gears or linkages, LAZA uses high-voltage electrostatic forces amplified by a liquid dielectric to directly actuate wings at the root, delivering specific power comparable to insect muscle (200 W/kg) \cite{helps2022liquid}. To delineate the lower functional limits of autonomous systems, it is instructive to examine platforms that push SWaP constraints to their extremes. Devices such as KickSat (5 g) \cite{manchester2013kicksat} and Wireless Capsule Endoscopes (7 g) \cite{swain2010remote} demonstrate computational autonomy at the gram scale. However, these platforms often necessitate reduced mobility or rely on passive locomotion, highlighting the severe trade-offs in actuation authority and control authority required as scale decreases below the nano-class \cite{neuman2022tiny}.

\subsection{Modular Ecosystems and the SWaP Penalty}
The usefulness of nano-scale research platforms like the Crazyflie is largely defined by their modular ecosystem of stackable PCBs, or ``decks''. Key examples include the Flow-deck v2, which provides optical flow and Time-of-Flight (ToF) ranging for non-GPS stability \cite{ostovar2022nano, 9381618}; the AI-deck, which integrates the GAP8 SoC for onboard ML inference \cite{palossi2021fully}; and the Multi-ranger, which utilizes arrays of VL53L1x sensors for omnidirectional distance sensing \cite{ostovar2022demo, arvidsson2023drone}. Additionally, the Lighthouse deck enables sub-millimeter external positioning using laser sweeps, offering a low-cost alternative to VICON systems \cite{green2019autonomous,bitcraze_lighthouse}. However, this modularity imposes a tangible penalty on flight performance. While the electronic components consume only a fraction of the power budget—adding an AI-deck increases electronic power consumption by only roughly 1.5\% \cite{palossi2021fully} (the primary cost is weight). The addition of a 4.4 g AI-deck reduces flight time by approximately 22\%, dropping endurance from $\sim$440 seconds to $\sim$340 seconds \cite{palossi2021fully, palossi201964}. Although individual sensors like $8 \times 8$ ToF arrays are lightweight, the cumulative weight of stacking multiple decks can quickly render the platform unflyable or drastically reduce agility \cite{ostovar2022nano}. Consequently, while the current state-of-the-art relies on modularity, future platforms will likely converge toward monolithic integration, consolidating the host MCU, AI accelerator, and memory into single-die or System-in-Package (SiP) solutions to maximize performance-per-watt and reclaim flight endurance.

\subsection{Sensing and Perception Architectures}
The sensing architecture of sub-50g nano-UAVs is also strictly governed by the same mentioned SWaP restrictions. With total system weights often capped at approximately 30 g and payload capacities limited to less than 15 g, the integration of perception sensors must not only satisfy minimal mass requirements but also operate within a power budget where the entire avionics suite (including sensing, computation, and control) is allotted merely 5--15\% of the total battery discharge, equating to approximately 100--200 mW \cite{valente2024heterogeneous, palossi2021fully, lamberti2024sim}. Consequently, sensor selection becomes a critical exercise in trading fidelity for efficiency.

\subsubsection{Visual Sensing Paradigms}
While larger micro-aerial vehicles may leverage Light Detection and Ranging (LiDAR) or high-fidelity stereo vision for SLAM, such modalities are generally prohibitive for nano-UAVs due to the mass and power density limits \cite{scarciglia2025map}. Consequently, monocular cameras have emerged as the dominant modality for this class of vehicles \cite{scarciglia2025map, lamberti2024sim}. The exclusion of stereo and RGB-D systems is dictated by both mechanical and computational constraints because the sub-10 cm frame diameter prevents the establishment of a sufficient baseline for accurate depth triangulation, while sensors like the RealSense exceed the entire weight budget of the platform \cite{scarciglia2025map, muller2023fully}. Even when miniaturised stereo vision systems are employed, the computational burden of disparity map estimation exceeds the limited instruction budget of onboard microcontrollers and consequently creates delays that undermine real-time control performance \cite{scarciglia2025map,crupi2024fusing}. As a result, many systems rely on lightweight CNNs or optical flow-based methods to infer depth and ego-motion \cite{conti2020memory}.

\subsubsection{Event-Based Neuromorphic Sensing}
To overcome the latency and redundancy limitations of standard frame-based cameras, recent research has pivoted toward bio-inspired Dynamic Vision Sensors (DVS). Unlike standard CMOS imagers that capture full frames synchronously, event cameras (DVS) operate asynchronously, transmitting data only when pixel log-luminosity changes exceed a set threshold \cite{bian2023colibriuav}. This paradigm generates a sparse stream of events rather than dense frames, drastically reducing data redundancy for static scenes while offering microsecond-level temporal resolution \cite{bian2023colibriuav}. A quintessential example is the iniVation AG DVS132S, which integrates a Synchronous Address Event Representation (SAER) readout interface. Unlike traditional USB interfaces that consume Watts of power, the SAER interface allows for a direct parallel connection to the host processor, supporting a throughput of up to 180 million events per second (Meps) while consuming only milliwatts \cite{bian2023colibriuav}. This asynchronous nature provides High Dynamic Range (HDR) capabilities and solves the ``data deluge'' problem, enabling perception-to-control latencies in the millisecond range vital for agile maneuvers \cite{bian2023colibriuav, 10935624}.

\subsubsection{Optical Flow and On-Chip Acceleration}
For velocity estimation, traditional optical flow calculation remains computationally expensive for general-purpose MCUs. To mitigate this, sensors such as the STMicroelectronics VD56G3 integrate a dedicated Application Specific Integrated Circuit (ASIC) directly on the sensor die \cite{kuhne2023fast}. This architecture offloads motion vector computation from the main MCU, achieving update rates of up to 300 fps at reduced resolutions while transmitting only flow vectors to conserve bandwidth \cite{kuhne2023fast}. Algorithmic efficiency is further maintained through sparse feature-based methods, such as FAST \cite{park2023development} corner detection and BRIEF descriptors, which limit tracking to a set number of vectors rather than dense pixel-wise estimation \cite{kuhne2023fast}. However, a critical failure mode involves motion blur induced by high-frequency motor vibrations; if exposure times are not minimized, the degradation of feature sharpness can lead to noisy or failed state estimation, necessitating strict mechanical damping or global shutter sensors \cite{park2023development}.

\subsubsection{Range Sensing}
The stringent SWaP constraints of nano-UAVs preclude the use of standard perception suites, such as commercial LiDAR units that weigh hundreds of grams and consume several watts \cite{muller2023robust, niculescu2022towards, zhou2022efficient}. Consequently, ToF sensors have emerged as the dominant ranging modality, offering direct metric depth data from milligram-scale packages \cite{niculescu2022towards, ostovar2022nano}. The technology has evolved from single-point sensors, like the STMicroelectronics VL53L1x used in multi-sensor arrays \cite{valente2024heterogeneous, friess2024fully, nguyen2021collision}, to multi-zone sensors such as the VL53L5CX, which provides a lightweight, 64-pixel matrix enabling basic obstacle avoidance and rough mapping via an I2C interface \cite{friess2024fully, niculescu2022towards, muller2023robust}. However, this reliance on miniaturized ToF sensors imposes significant perceptual limitations compared to larger platforms. Single-beam sensors feature a narrow Field of View (FoV), creating ``tunnel vision'' and vulnerability to narrow obstacles \cite{muller2024batdeck, lamberti2024combining}. While multi-zone sensors improve the FoV, they still fail to match the omnidirectional coverage of scanning LiDARs, often requiring multiple, angled sensors that add integration complexity \cite{friess2024fully, muller2023robust, crupi2025efficient}. Furthermore, the low resolution (e.g., 64 pixels) and coarse angular granularity result in ``mixed pixels'' and increasing position uncertainty with distance \cite{niculescu2022towards, niculescu2023nanoslam}. Effective range is typically limited to $<4$ m, with data validity degrading significantly beyond $\sim$2 m, forcing nano-UAVs to operate with reduced safety margins \cite{ostovar2022demo, muller2023robust, niculescu2022towards, niculescu2023nanoslam}.

\subsubsection{Internal State and Signal Sensing}
Robust autonomy for nano-UAVs relies on precise internal state estimation and non-visual localization, which must balance sensor fidelity against the computational overhead of data fusion under strict SWaP constraints \cite{greiff2017modelling}. Inertial measurement is typically anchored by MEMS sensors like the InvenSense MPU-9250. However, as these share a rigid frame with high-RPM motors, the raw data is heavily corrupted by vibration noise, necessitating rigorous software-based low-pass filtering that consumes valuable microcontroller cycles before state estimation begins \cite{greiff2017modelling}. The fusion of this noisy inertial data with external updates presents a critical computational trade-off. While lightweight complementary filters (e.g., Madgwick) are efficient, they suffer from yaw drift due to a lack of predictive modeling \cite{greiff2017modelling}. In contrast, Extended Kalman Filters (EKF) enable robust fusion of asynchronous measurements (e.g., from optical flow or UWB) by explicitly modelling sensor noise, but their matrix operations demand a significant portion of the STM32's instruction cycles, potentially starving higher-level tasks \cite{greiff2017modelling}.

To augment inertial navigation in GPS-denied environments, nano-UAVs leverage radio and acoustic modalities. Received Signal Strength Indicator (RSSI) methods are highly SWaP-efficient, utilizing existing communication hardware, but are unsuitable for precise metric localization due to multipath effects and noise \cite{van2020board, mcguire2019minimal}. Ultra-Wideband (UWB) provides centimeter-level accuracy via ToF ranging, yet incurs a severe power penalty—consuming up to 480 mW and constituting over 30\% of the non-actuator power budget—while added mass reduces flight endurance \cite{niculescu2022energy, pourjabar2023land, lucahigh}. Emerging acoustic sensing research exploits ``ego-noise'' for echolocation to detect surfaces impervious to optical sensors. However, this requires sophisticated signal processing to isolate environmental reflections from overwhelming self-generated motor noise, adding complexity to the processing pipeline \cite{friess2024fully,muller2024batdeck}.

Ultimately, the sensory apparatus of sub-50g nano-UAVs is defined by a ``Sensing Gap'': a fundamental disparity between the rich, high-fidelity data required for safe autonomy and the sparse, noisy, low-resolution data available. Forced to rely on monocular cameras lacking depth \cite{scarciglia2025map}, sparse ToF sensors with limited range \cite{muller2023robust, greiff2017modelling}, and vibration-corrupted IMUs \cite{greiff2017modelling}, the burden of closing this gap shifts entirely to software. The algorithmic stack must not only control vehicle dynamics but also compensate for missing state information and filter aggressive noise, necessitating the highly optimized and often learning-based control strategies discussed in the subsequent section.

\section{Algorithms and Control Architectures}
\label{sec:algorithms}

\subsection{The Shift to Learning-Based Perception}
\begin{figure*}[t!]
    \centering
    \includegraphics[width=0.85\textwidth]{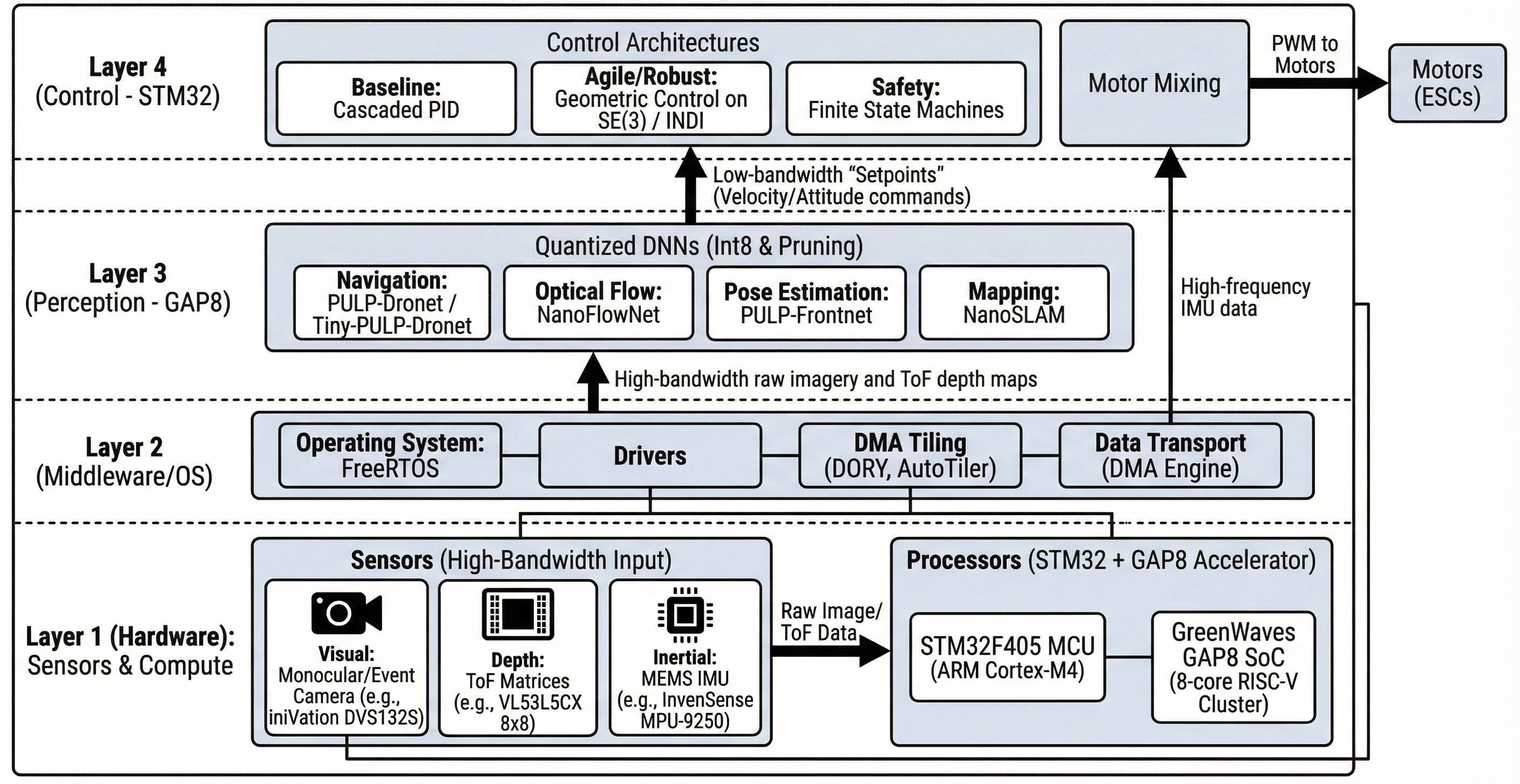}
    \caption{\textbf{The ``Edge-AI'' Autonomy Stack.} A hierarchical view of the hardware-software co-design for nano-UAVs. High-bandwidth sensor data flows from Layer 1 into the perception engine (Layer 3) on the AI accelerator (GAP8), which generates compressed state estimates for the real-time control loop (Layer 4) on the MCU (STM32). Middleware (Layer 2) manages the critical data transport between these heterogeneous cores.}
    \label{fig:autonomy_stack}
\end{figure*}

The migration of autonomous navigation logic from classical, geometry-based methods to deep learning based methods represents a fundamental shift in the design of nano-UAVs, driven primarily by the severe SWaP constraints inherent to the platforms. While classical SLAM and VIO provide mathematical guarantees and precision, they often incur prohibitive computational and memory costs for sub-50g vehicles \cite{lamberti2024combining}. For instance, maintaining dense metric maps requires memory resources that far exceed the typical (less than 1MB) SRAM available on standard microcontrollers like the STM32F405 \cite{niculescu2023nanoslam, akbari2024tiny}. Furthermore, classical estimators are brittle when subjected to the high-frequency vibrations and abrupt lighting changes characteristic of nano-drone operations \cite{conti2020memory}. In contrast, deep learning models have demonstrated superior robustness to sensory noise and dynamic environments by learning high-level abstractions directly from raw sensor data, enabling navigation without the necessity of explicit, memory-intensive map maintenance \cite{lamberti2024combining, cereda2023secure}.To manage the complexity of onboard processing, the software architecture is organized into a hierarchical ``Autonomy Stack'' (see Fig. \ref{fig:autonomy_stack}). This structure layers high-bandwidth perception tasks on parallel accelerators over real-time control loops running on the main MCU, ensuring that computationally intensive inferences do not block critical flight stabilization functions.
\subsubsection{CNN Architectures for Edge Inference}
The deployment of Convolutional Neural Networks (CNNs) on sub-50g nano-UAVs has evolved from direct porting of standard architectures to the development of highly specialized, hardware-aware models. This evolution is best exemplified by the trajectory of the DroNet \cite{eschmann2024learning} family. The original DroNet architecture, a bifurcated CNN inspired by ResNet-8 \cite{he2016deep}, processed monocular images to produce steering angles and collision probabilities \cite{lamberti2024sim, panerati2021learning}. However, it relied on powerful external processors consuming tens of Watts \cite{palossi201964} (two orders of magnitude higher than the $<$100 mW budget available on a nano-UAV) rendering it infeasible for onboard execution \cite{palossi201964, palossi2021fully}. To bridge this gap, PULP-Dronet \cite{lamberti2022tiny} was introduced as the first successful port of the architecture to the GreenWaves GAP8 RISC-V SoC \cite{palossi201964}. The core innovation was the abandonment of the sequential processing paradigm in favor of Parallel Ultra-Low Power (PULP) computing. By utilizing the GAP8’s 8-core cluster and quantizing weights from 32-bit floating-point to 16-bit fixed-point (to circumvent the lack of an FPU), the system achieved inference rates of 6--18 frames per second (fps) within a power envelope of 64--272 mW \cite{palossi201964}. Despite these optimizations, the reliance on low-resolution imagery necessitated fine-tuning on domain-specific datasets to maintain safe flight capabilities up to 1.5 m/s \cite{palossi2019open}. The subsequent iteration, Tiny-PULP-Dronet (and related variants like the ``Generalization through Simulation'' framework \cite{lamberti2024sim}), pushed these optimizations to the extreme to maximize throughput and minimize the ``Sim-to-Real'' gap. By aggressively quantizing to 8-bit integers and pruning network parameters, computational cost was reduced to approximately $\sim$1 MMAC/frame, enabling peak throughputs of 135--160 fps \cite{lamberti2024sim, aydinli2023deep}. In collision avoidance tasks, this high-speed inference resulted in a 25.3\% improvement in the speed/braking-distance ratio compared to the original PULP-Dronet \cite{niculescu2021improving}. However, this aggressive reduction in model capacity poses a risk of overfitting to specific visual features of the training environment, potentially reducing generalization in visually distinct unseen environments.

Beyond general navigation, the field has diversified into task-specific architectures optimized for the edge. For optical flow estimation, NanoFlowNet \cite{bouwmeester2022nanoflownet} was developed to provide dense, per-pixel flow estimation on the GAP8 SoC, replacing the computationally prohibitive FlowNet2. By replacing regular convolutions with depth-wise separable convolutions and utilizing ``motion boundary detail guidance'' during training, NanoFlowNet achieves inference rates between 5.5 and 9.3 fps. While this is lower than desktop-class GPUs, it outperforms squeezed variants of standard networks that fail to fit within the GAP8’s memory constraints. For Human-Drone Interaction (HDI), PULP-Frontnet adapts the ResNet architecture for relative pose estimation. By reducing input resolution and utilizing 8-bit quantization, the model fits within the L2 memory of the GAP8 SoC, scaling from 6 fps (64 mW) to a peak throughput of 135 fps (86 mW) \cite{palossi201964, palossi2021fully}. However, this extreme resolution reduction introduces a measurable drop in regression accuracy for orientation parameters, highlighting the hard limit of SWaP-constrained computer vision where geometric precision is sacrificed for computational throughput. Similarly, for drone-to-drone localization, Fully CNNs (FCNNs) have been proposed that output probability maps rather than scalar coordinates \cite{lucahigh, crupi5352666blinking}. Running at 39 Hz (101 mW), these models significantly outperform standard baselines like MobileNetV2 in tracking error while simultaneously classifying target LED states for visible light communication \cite{crupi5352666blinking}.

\subsubsection{Optimization for the Edge: Quantization, Pruning, and Auto-Tiling}
The deployment of CNNs on sub-50,g nano-UAVs is fundamentally limited by the memory hierarchy and energy efficiency of the onboard computing substrate. Standard deep architectures (e.g., ResNet-50) demand computational resources that exceed the $<100$~mW power budget of nano-drones by orders of magnitude \cite{palossi201964}. To reconcile algorithmic complexity with strict SWaP constraints, the literature applies three complementary optimisation strategies: quantization, pruning (architectural sparsity), and automated memory tiling.

Reducing numeric precision is dictated both by algorithmic tolerance and by the target edge architecture. On processors without hardware FPUs, such as the GreenWaves GAP8 cluster, executing a float32 model requires software emulation, which incurs large execution overheads and precludes real-time control loops \cite{lucahigh}. Early edge implementations adopted 16-bit fixed-point formats (e.g., Fixed16 Q4.12), which provided sufficient dynamic range for weights and activations while avoiding accumulator saturation. This representation enabled the GAP8 to run navigation networks at up to $\sim$18,fps within a modest power envelope \cite{palossi201964}. More recently, aggressive 8-bit integer quantisation (Int8) has become the de facto standard on many platforms. Transitioning from 16-bit to 8-bit representations reduces memory footprint substantially and increases inference throughput by exploiting SIMD instructions on modern RISC-V cores; empirical studies report large throughput gains with minimal degradation in regression and classification performance \cite{scarciglia2025map, crupi2025efficient, lucahigh, akbari2024tiny, niculescu2021improving}.

A critical distinction in this domain is the choice between Post-Training Quantization (PTQ) and Quantization-Aware Training (QAT).  PTQ offers rapid deployment by calibrating pre-trained weights using a representative dataset, whereas QAT simulates quantization noise (e.g., rounding errors) during the training backward pass, forcing the network to adapt to lower precision at the cost of complex retraining. The evolution of the PULP-Dronet architecture illustrates the shifting preference between these paradigms as toolchains mature. The original implementation (V1) necessitated QAT to stabilize its 16-bit fixed-point control policy on the GAP8 SoC without accuracy loss \cite{palossi201964,palossi2019open}. However, subsequent automated iterations (V2) transitioned to aggressive 8-bit PTQ leveraging the NEMO/DORY toolchain \cite{lamberti2022tiny, niculescu2021improving}. This shift demonstrates that modern deployment flows can now preserve feature fidelity through advanced calibration statistics alone, eliminating the prohibitive computational overhead of retraining required by earlier architectures.

Where quantization reduces parameter precision, pruning reduces parameter count. For embedded targets, structured pruning is typically preferred because it removes entire channels or filters and therefore maps efficiently to vectorised hardware primitives, unlike unstructured sparsity which hampers memory access patterns and accelerator utilisation \cite{risso2024optimized}. Automated methods, including neural architecture search variants and cascade pruning algorithms, have been shown to identify compact architectures at iso-accuracy, yielding substantial model size reductions and latency improvements. For instance, cascaded Supernet/PIT strategies discovered MobileNet-derived configurations that achieve $\sim$25\% size compression at equal accuracy or reduce inference latency by severalfold; extreme compression variants (e.g., Tiny-PULP-Dronet) lower computational complexity to the order of 1 MMAC per frame to enable very high inference rates \cite{risso2024optimized, scarciglia2025map}.

Effective deployment also requires explicit management of the constrained memory hierarchy of edge silicon. Many embedded processors (e.g., GAP8) provide small, explicitly managed scratchpad memories (e.g., 64 kB L1, 512 kB L2) rather than large hardware caches \cite{9381618}. Naïve libraries often fail to orchestrate data movement between these levels efficiently, causing pipeline stalls. Specialized toolchains therefore model memory management as an optimisation problem and automatically generate tiled code that fits working sets into L1. Tools such as DORY and the GreenWaves AutoTiler decompose network layers into tiles, schedule DMA transfers, and employ software pipelining with double buffering so that while cluster cores compute on the current tile, the DMA engine prefetches the next tile from L2 \cite{9381618, niculescu2021improving, crupi2025efficient}. This overlap of computation and communication hides memory latency and keeps execution units saturated, producing significant throughput improvements over vendor supplied baselines on representative convolution kernels \cite{9381618}.

\subsubsection{The Training Pipeline: Sim-to-Real Transfer and Adaptation}
\begin{figure*}[t!]
    \centering
    \includegraphics[width=0.95\textwidth]{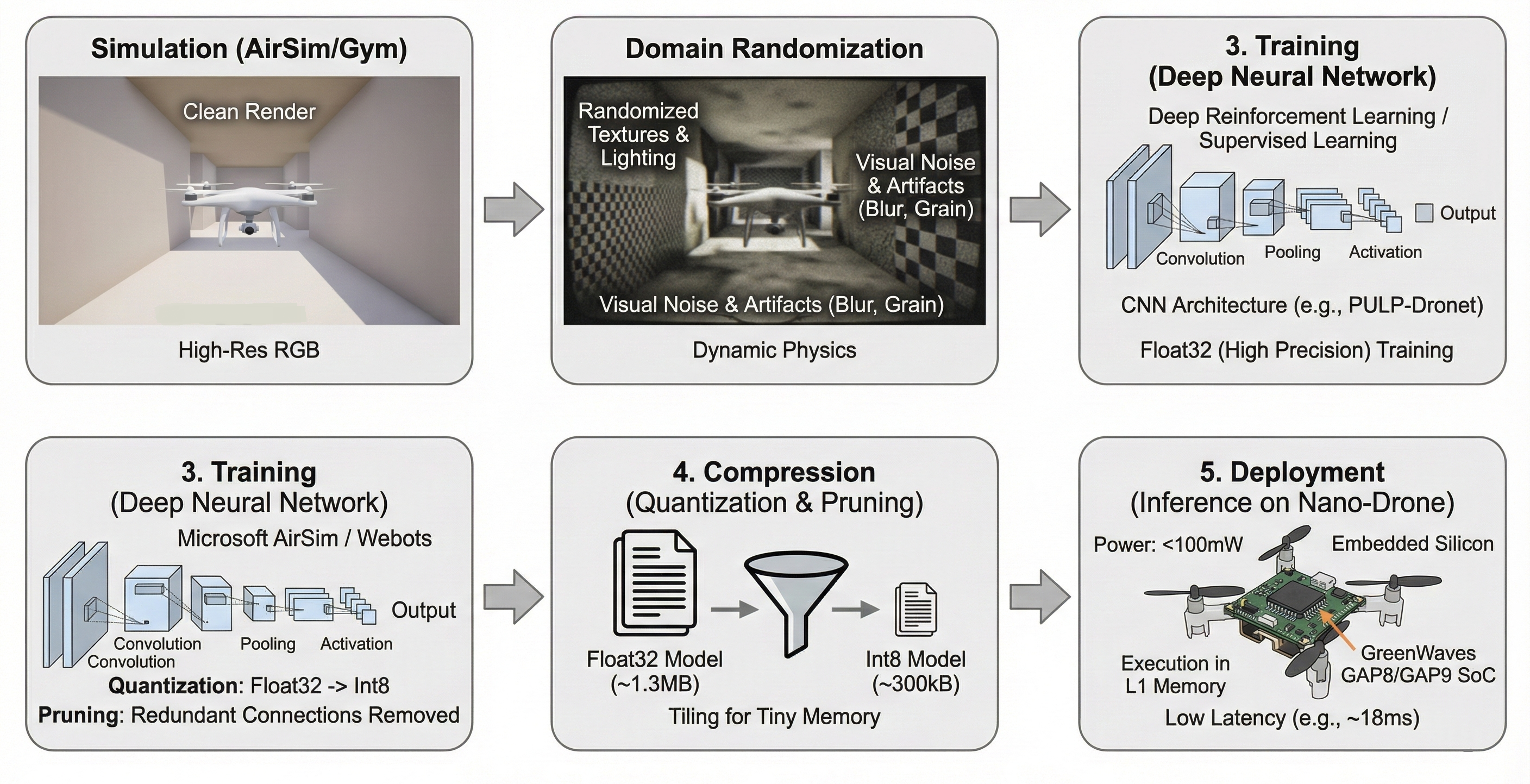}
    \caption{\textbf{The Sim-to-Real Training Pipeline.} A five-step workflow bridging the gap between simulation and reality. (1) Training begins in high-fidelity simulators (e.g., AirSim). (2) Domain Randomization injects visual noise to induce robustness. (3) Policies are trained using Deep RL or Supervised Learning. (4) Models undergo aggressive compression (Int8 quantization, pruning). (5) The optimized binary is deployed to the nano-UAV's limited L1 memory for real-time inference.}
    \label{fig:sim_to_real}
\end{figure*}
The deployment of DNNs on nano-UAVs is fundamentally restricted by the ``Sim-to-Real'' gap--the discrepancy between the synthetic environments used for safe, massive-scale training and the stochastic, noisy physical world. As illustrated in Fig. \ref{fig:sim_to_real}, successfully bridging this gap necessitates a rigorous five-step pipeline that decouples heavy offline training from lightweight online inference. The process transitions from high-fidelity simulation with domain randomization to aggressive model compression (quantization and pruning) before final deployment on the target edge hardware.
 While standard-sized UAVs may support heavy, redundant sensor suites to correct model mismatches, nano-UAVs operating within sub-100mW power envelopes lack the payload for such corrective hardware. Consequently, the burden of robustness shifts entirely to the training methodology, necessitating pipelines that decouple heavy offline learning from lightweight online inference.

The primary obstacle to training Reinforcement Learning (RL) or imitation learning policies directly on physical nano-UAVs is hardware fragility. Unlike ground robots, nano-quadrotors cannot safely ``fail'' during the exploration phase of RL; a single collision often terminates data collection or damages the platform \cite{eschmann2024learning,liu2022adaptive}. Furthermore, high-fidelity simulation of nano-scale aerodynamics is notoriously difficult due to complex nonlinear effects such as rotor drag and ground effects \cite{nguyen2021collision, lambert2019low}. To bridge this reality gap without dangerous physical interaction, domain randomization has emerged as the standard methodology for policy synthesis. This technique exposes the agent to a vast spectrum of simulated environments during training, randomizing parameters such as lighting, texture, and physical dynamics to force the DNN to learn invariant features \cite{kang2019generalization}. For vision-based navigation, techniques such as ``Generalization through Simulation'' randomize visual textures to prevent overfitting to synthetic artifacts \cite{kang2019generalization}. Recent advancements include background randomization, where the background of training images is synthetically replaced to force the network to focus solely on the subject (e.g., for human pose estimation), improving regression performance by up to 40\% in unseen real-world environments \cite{cereda2021improving, stolfi2024argos}. Beyond visuals, dynamics randomization varies physical parameters such as mass, motor thrust coefficients and drag to ensure policies remain robust against the manufacturing variability inherent in off-the-shelf nano-drone components \cite{huang2023dattdeepadaptivetrajectory,kang2019generalization}.

A critical, often overlooked aspect of the Sim-to-Real pipeline is the accurate modelling of low-power sensor imperfections. The low-resolution (QVGA), monochrome cameras used on platforms like the Crazyflie suffer from significant noise, motion blur, and vignetting. Successful augmentation pipelines must inject Gaussian noise, synthetic blur, and random gamma corrections during training to produce networks that are robust to the degraded image quality encountered during agile flight \cite{palossi2021fully, cereda2021improving}. This ensures that the static, compressed neural network deployed on the MCU is resilient to sensory noise without requiring additional onboard computation. While domain randomization produces robust static policies, it does not allow agents to adapt to novel concepts encountered post-deployment. To address this, federated continual learning has been proposed as a paradigm for collective learning within swarms \cite{kroger2025device}. In this architecture, individual drones learn from local data streams and share model updates (gradients or weights) rather than raw data, thereby circumventing the prohibitive energy and bandwidth costs of centralized video transmission. By performing training locally on AI-capable SoCs (e.g., GAP9), this strategy was demonstrated to require only 24 kB of data transmission per node, completing a global training epoch with a total energy cost of merely 4.3 mJ per local epoch. Furthermore, the integration of mean output loss regularization allows these nano-drones to learn new classes incrementally within strict memory limits (e.g., 1.5 MB RAM) without suffering from catastrophic forgetting.

\subsection{Navigation and Trajectory Generation}
\label{sec:navigation}

Navigation architectures for nano-UAVs generally fall into two distinct paradigms: Map-based (Modular) and Map-less (End-to-End).  Map-based systems, such as NanoSLAM \cite{niculescu2023nanoslam}, decouple the autonomy stack into explicit mapping, planning, and control modules. This offers interpretability and safety guarantees through metric representations (e.g., occupancy grids) but incurs significant memory overhead to maintain state \cite{niculescu2023nanoslam}. Conversely, Map-less approaches, exemplified by PULP-Dronet \cite{palossi201964}, utilize end-to-end learning to map raw sensor inputs directly to control actuation via neural networks. While this minimizes latency and memory footprint (ideal for SWaP-constrained nano-UAVs) it lacks the long-horizon planning capabilities and interpretability of modular stacks \cite{palossi201964}. The following sections analyse how nano-UAVs navigate these trade-offs.

\subsubsection{Global Planning and Exploration: Bridging the Memory Gap}
The ability to plan long-horizon trajectories in unknown environments is the cornerstone of autonomous exploration. However, on sub-50g nano-UAVs, this capability is fundamentally restricted by the scarcity of onboard Static Random Access Memory (SRAM), typically limited to less than 1 MB (e.g., 192 kB on the STM32F405 or 512 kB on the GAP8 SoC). Unlike standard-class UAVs that utilize gigabytes of memory to maintain dense euclidean signed distance fields or voxel maps, nano-drones face a ``Mapping Gap'': they must represent the environment with sufficient fidelity for safe navigation while ensuring the map footprint does not displace the control stack. Consequently, the field has bifurcated into lightweight occupancy grids for reactive exploration and hierarchical graph-based methods for metric accuracy.

To circumvent the memory limitations of metric mapping, researchers have adapted the tinySLAM algorithm for nano-platforms. Originally designed for laser range finders, tinySLAM utilizes a discretized occupancy grid where cells store 8-bit integers representing confidence values (0--255) rather than floating-point probabilities \cite{markdahl2023tinyslam}. On the Crazyflie 2.1, a cell map at 10 cm resolution occupies only 10 kB of the available 192 kB RAM. To facilitate convergence with sparse ToF sensors (such as the $8 \times 8$ pixel VL53L5CX) the algorithm employs a ``hole function'' that blurs measured obstacles into neighbouring cells, effectively thickening walls to compensate for sensor noise and sparsity. For higher fidelity, Graph-Based SLAM (Pose Graph Optimization) is often preferred over memory-intensive particle filters like RBPF, which suffer from particle depletion in large environments. However, solving the linear system for graph optimization can bottleneck the MCU. NanoSLAM \cite{niculescu2023nanoslam} addresses this via a hierarchical optimization strategy: the full pose graph is divided into subgraphs, and only a sparse graph of keyframes is optimized globally. This approach constrains the linear solver’s memory footprint to fit within the 128 kB L1 memory of the GAP9 cluster while correcting trajectory errors by up to 67\% compared to dead reckoning.

Once a map is constructed, the planner must generate a collision-free path to the goal. Classical global search algorithms, such as A*, provide resolution-complete pathfinding but suffer from the ``curse of dimensionality'' when applied to high-fidelity grids \cite{samavedula2025mini}. To maintain real-time performance in 3D cluttered environments, Jump Point Search (JPS) is utilized to prune symmetric path segments, significantly reducing the number of nodes expanded \cite{toumieh2024high}. However, a critical limitation of discrete planners (A*/JPS) is that they generate geometrically feasible but dynamically infeasible paths (composed of straight lines and sharp turns that defy the kinematic limits of high-speed nano-quadrotors) \cite{toumieh2024high, greiff2017modelling}. To resolve this, modern frameworks like High-Speed Decentralized Synchronous Motion (HDSM) decouple global planning from local trajectory generation. HDSM employs JPS on an inflated voxel grid to identify a geometric skeleton ($T_{path}$), which then defines a spatially restricted ``search corridor'' \cite{toumieh2024high}. This corridor effectively prunes the solution space for downstream trajectory optimizers ensuring that computationally expensive operations like model predictive control are only evaluated within a narrow volume of valid airspace rather than the entire map. Underpinning both mapping and planning is the requirement for robust odometry. While standard VIO pipelines are computationally prohibitive for general-purpose MCUs, specialized hardware acceleration is emerging as a viable solution. The Navion chip, a 24 mW ASIC, performs full VIO integration on-die, utilizing lossy image compression and structured sparsity to reduce memory requirements by 4.1 times \cite{suleiman2019navion}. Alternatively, software-based solutions on the GAP8 SoC achieve velocity estimation by fusing optical flow with ToF height measurements, although these systems remain vulnerable to scale ambiguity and drift if not corrected by loop closures within the SLAM backend \cite{palossi2021fully, greiff2017modelling}.

\subsubsection{Local Planning and Obstacle Avoidance}
While global planning ensures long-term mission success, the immediate safety of nano-UAVs in cluttered, GPS-denied environments depends on rapid, local obstacle avoidance. Given the strict SWaP constraints nano-drones frequently lack the computational capacity for continuous trajectory optimization. Consequently, the field heavily utilizes reactive heuristics and bio-inspired reflexes that map raw sensor data directly to control actions, bypassing the need for heavy metric mapping.

Heuristic approaches prioritize low computational latency and minimal memory footprints, making them the \textit{de facto} standard for microcontroller-scale autonomy. These methods typically employ finite state machines that switch behavioral modes based on immediate sensor thresholds. A quintessential example is the Swarm Gradient Bug Algorithm (SGBA), designed for the 27 g Crazyflie 2.0 \cite{mcguire2019minimal}. SGBA integrates gradient ascent (for source seeking) with a ``wall-following'' behavior for obstacle avoidance using only four single-pixel laser rangers. Instead of building a memory-intensive map, the drone reacts to obstacle detection by tracing the boundary of the object until its direct path to the goal is clear. This approach is exceptionally efficient, requiring negligible memory (fitting comfortably within 196 kB RAM) and allowing large swarms to navigate complex environments without inter-agent communication. However, such purely reactive heuristics are susceptible to local minima, potentially leading to infinite loops in complex geometries if loop-detection logic is not implemented. For platforms equipped with monocular cameras but lacking the computational budget for dense depth estimation, reactive planning can be derived directly from 2D object detection outputs. Sartori et al. \cite{sartori2025ai, sartori2023autonomous} proposed a heuristic where ``collision risk'' is approximated by the width of an obstacle's bounding box in the camera frame. A repulsive velocity vector is computed to steer the drone toward the image half-plane containing more free space. While this method bypasses the latency of stereo processing or depth prediction, it relies heavily on the reliability of the upstream detector. False negatives (failure to detect an object) can lead to immediate collisions, and the lack of metric depth necessitates conservative tuning of repulsion gains, often resulting in suboptimal path efficiency.

Biological systems, such as honeybees, navigate dense clutter without metric maps by exploiting optical flow fields. NanoFlowNet adapts this paradigm for nano-UAVs by implementing a horizontal balance strategy \cite{bouwmeester2022nanoflownet}. The control law adjusts the drone's yaw rate to equalize the magnitude of optical flow observed on the left and right sides of the FoV. This simple mechanism generates an emergent centering behaviour in corridors and provides natural obstacle avoidance. To address the ``focus-of-expansion'' problem (where flow is minimal during head-on approaches or generally low in textureless environments) the drone may induce active oscillations (up-down movement) to generate observable flow. While this method is computationally efficient enough to execute on the GAP8 SoC at approximately 9 FPS, it remains sensitive to environmental texture and lighting conditions, unlike active sensing methods (e.g., laser ranging). While finite state machines provide computational efficiency, they often result in ``jerky,'' discrete state switches (e.g., switching abruptly from ``forward'' to ``stop'') that induce mechanical stress and consume excess power. To address this, physics-based approaches model the drone as a particle within a potential or fluid field, generating smooth, continuous trajectories. The fluid flow navigation framework \cite{uppaluru2023multi} treats cooperative agents as particles sliding along streamlines that mathematically wrap around obstacles modelled as singularities. This formulation ensures that avoidance manoeuvres are continuous and differentiable, preventing the harsh kinematic discontinuities seen in bug algorithms. Similarly, acceleration fields \cite{gonccalves2024safe} compute target acceleration vectors that simultaneously guide agents toward goal positions while repelling them from obstacles and peers. By modelling obstacles as repulsive cylinders and generating smooth avoidance vector fields, these methods mitigate the abrupt braking manoeuvres characteristic of purely reactive finite state machines, thereby reducing mechanical wear and extending flight endurance.

\subsubsection{Optimization-Based Local Planning and Safety Filters}
While heuristics provide low-latency reactivity, they lack the capacity to handle dynamic constraints or guarantee safety. To address this, the field has gravitated toward model predictive control (MPC) and formal safety filters. However, standard nonlinear MPC (NMPC) is computationally prohibitive for nano-drones; solving non-convex optimization problems at control rates ($>$50 Hz) exceeds the floating-point capability of MCUs like the STM32F4. Consequently, research has focused on algorithmic distillation and convex relaxation to achieve edge feasibility.

Research has prioritized adapting MPC specifically for constrained hardware by reducing solver complexity. Sampling-based methods, such as Model Predictive Path Integral (MPPI) control, offer robustness against non-smooth cost functions by sampling thousands of potential trajectories. However, standard MPPI implementations typically mandate GPU acceleration (e.g., NVIDIA Jetson) to achieve real-time performance, rendering them unsuitable for microcontroller-based nano-drones \cite{pravitra2020ℒ}. To mitigate this, deep model predictive optimization has been proposed to learn the inner-loop optimizer of MPPI, achieving superior performance with fewer samples and reduced memory requirements \cite{10611492}. Furthermore, to bridge the ``Sim-to-Real'' gap, L1-Adaptive control has been integrated with MPPI to compensate for both matched and unmatched uncertainties (such as aerodynamic drag) thereby recovering robustness without the heavy computational cost of full NMPC \cite{pravitra2020ℒ}. Perhaps the most significant advancement for nano-UAVs is Tiny LB MPC \cite{akbari2024tiny}, which explicitly targets the constraints of microcontrollers like the Teensy 4.0 (compatible with the Crazyflie 2.1). By exploiting the differential flatness of multirotor dynamics, this framework transforms the nonlinear optimal control problem into a convex second-order cone program. Crucially, it employs a custom Alternating Direction Method of Multipliers (ADMM) solver where expensive primal updates are pre-computed at a lower rate (10 Hz) using a linearized Gaussian Process. This architectural splitting allows the controller to execute at 100 Hz, outperforming standard flatness-based MPC by 23\% in tracking error while remaining computationally feasible for edge deployment \cite{akbari2024tiny}. 

While optimization improves trajectory quality, ensuring safety without over-conservatism remains critical. Control Barrier Functions (CBFs) and Model Predictive Safety Filters (MPSFs) provide mathematical guarantees of forward invariance of the safe set (i.e., crash-free operation). The primary challenge lies in the computational cost. Filtering unsafe control inputs typically requires solving a Quadratic Program (QP) at every control step \cite{tayal2024polygonal}, which can saturate an onboard MCU running at 50--100Hz. Standard collision cone CBFs often overestimate obstacle size by treating them as circles, which wastes navigable space in cluttered environments. To address this, Polygonal Cone CBFs (PolyC2BF) construct cones based on the vertices of polygonal obstacles. This method formulates safety as a Lipschitz continuous QP, allowing for real-time execution in tight spaces where circular approximations would falsely detect a collision \cite{tayal2024polygonal}.

Recent advancements suggest that formal methods are becoming increasingly edge-feasible through algorithmic restructuring. For instance, AMSwarm \cite{adajania2023amswarm} utilizes alternating minimization to solve trajectory optimization problems with collision constraints. By reformulating constraints into a polar form, it avoids the conservative linearization required by Sequential Convex Programming (SCP) and retains a QP structure solvable on embedded MCUs. This approach has demonstrated success rates 72\% higher than SCP baselines while running in approximately 4.4 ms per agent \cite{adajania2023amswarm}. Furthermore, multi-step MPSFs implemented on the Crazyflie 2.0 have reduced control ``chattering'' by 78\% compared to single-step filters, proving that predictive safety filtering is viable on STM32-class processors if the solver is highly optimized \cite{bejarano2023multi}. Ultimately, the feasibility of optimization-based planning on nano-UAVs relies on specialized solvers. General-purpose solvers like OSQP do not fully exploit the structure of the specific control problem. Custom implementations, such as the ADMM splitting in Tiny LB MPC or the alternating minimization in AMSwarm, are required to accommodate the limited floating-point performance of nano-scale processors. While full NMPC remains currently infeasible for onboard execution without hardware acceleration, these distilled approaches represent the state-of-the-art in delivering high-performance control within the $<$100 mW budget.

\subsubsection{RL-Based Navigation: Learning Policies for Cluttered Environments}
While heuristics offer computational efficiency, they often fail to capture the coupled dynamics required for aggressive manoeuvres or robust collision avoidance in highly stochastic environments. Consequently, Reinforcement Learning (RL) has emerged as a dominant paradigm for navigating complex spaces where explicit modelling of sensor noise and aerodynamic disturbances is intractable. The deployment of RL on nano-UAVs is strictly governed by the ``Inference Asymmetry'' principle: while training requires vast computational resources to solve the Bellman equation, the resulting policy is typically a static, lightweight neural network (e.g., a Multi-Layer Perceptron with two hidden layers of 64 neurons) \cite{kooi2021inclined, eschmann2024learning}. This architecture aligns perfectly with the sub-100mW constraints of the GAP8 or STM32 microcontrollers, enabling high-frequency control loops (50--100 Hz) that would be choked by the iterative optimization required in model-based approaches like Nonlinear MPC \cite{lambert2019low, eschmann2024learning}. Among continuous control algorithms, Proximal Policy Optimization (PPO) has become the standard for nano-UAVs \cite{kooi2021inclined}. Although theoretically less sample-efficient than off-policy methods like Soft Actor-Critic (SAC) or Twin Delayed DDPG (TD3), empirical evaluations on platforms like the Crazyflie 2.1 indicate that PPO offers superior stability for tasks requiring curriculum learning. For instance, in complex terminal guidance tasks such as landing on inclined surfaces, PPO successfully converges where replay-buffer-based methods (SAC/TD3) struggle to adapt to the changing goal difficulty due to buffer staleness.

A critical barrier to robust RL-based navigation is the severe ``Partial Observability'' inherent to nano-UAVs, which lack high-fidelity state estimators like LiDAR or VICON. To address this, Asymmetric Actor-Critic architectures decouple the information available during training from that available during deployment \cite{eschmann2024learning}. In this paradigm, the ``Critic'' network is trained in simulation using privileged ground-truth information—such as exact obstacle positions, wind vectors, and inertial states—while the ``Actor'' (the policy deployed on the drone) is constrained to use only the noisy, partial histories from onboard sensors like the IMU and low-resolution depth imagers. This structural asymmetry allows the deployed policy to implicitly learn robust representations of unmeasured dynamics (e.g., rotor drag or wind gusts) without the computational penalty of running a state estimator. Ablation studies demonstrate that removing this privileged training information significantly degrades tracking accuracy, as the policy fails to compensate for unobserved latent states.

Standard RL algorithms typically maximize expected returns, often resulting in ``risk-neutral'' behaviours that are dangerous for fragile nano-drones operating in stochastic environments. To mitigate collision risks without sacrificing navigational efficiency, Adaptive Risk-Tendency Implicit Quantile Network (ART-IQN) introduces a distributional RL approach \cite{liu2022adaptive}. Unlike standard methods that estimate a single value expectation, ART-IQN estimates the distribution of future returns, utilizing the Lower Tail Conditional Variance as a metric for intrinsic environmental uncertainty. By forecasting this uncertainty, the algorithm dynamically adjusts the agent's risk tendency (Conditional Value at Risk, or CVaR) in real-time. Consequently, a nano-drone equipped with simple laser rangers can autonomously exhibit adaptive behaviour: navigating aggressively in open space to maximize speed while switching to conservative, risk-averse flight in cluttered areas, achieving higher success rates than static baseline policies.

\subsection{Flight Control and Dynamics}
The flight control architecture of nano-UAVs is strictly governed by the necessity to stabilize an inherently unstable, underactuated system with extremely low inertia. Unlike standard-sized quadrotors, where mass provides passive rejection of high-frequency disturbances, nano-UAVs (e.g., the 27 g Crazyflie 2.1) exhibit rapid dynamics that require high-bandwidth control loops (typically $>500$ Hz) \cite{green2019autonomous, eschmann2024learning}. Given the sub-100 mW power envelope of onboard microcontrollers (e.g., STM32F405), the selection of a control law is a strict trade-off between algorithmic complexity and the ability to handle the full flight envelope.

\subsubsection{Classical and Geometric Control Architectures}

The baseline control strategy for nano-UAVs remains the cascaded PID architecture. This approach decouples the system into an outer position loop (operating at $\approx$100 Hz) and an inner attitude/rate loop (operating at $\approx$500 Hz) \cite{luis2016design, de2024testing}. Due to its $\mathcal{O}(1)$ computational complexity, PID control executes efficiently on low-power Cortex-M4 microcontrollers, occupying minimal CPU cycles \cite{giernacki2017crazyflie}. Similarly, linear quadratic regulator (LQR) offers theoretically optimal control by minimizing a quadratic cost function of state error and control effort. While LQR outperforms PID in station-keeping tasks by explicitly accounting for state coupling, it fundamentally relies on a linearized model of dynamics around a hover equilibrium \cite{de2011modeling}. Critically, both PID and time-invariant LQR suffer from fundamental limitations during aggressive manoeuvres. These linear controllers assume small angle approximations (e.g., $\sin(\phi) \approx \phi$). Consequently, when a nano-drone attempts manoeuvres involving large roll/pitch angles ($>30^\circ$) or inverted flight, the linearization breaks down, and Euler-angle singularities (Gimbal lock) can cause catastrophic loss of control \cite{greiff2017modelling}. Recent work has sought to clone the behaviour of these robust linear controllers into more efficient substrates. Stroobants et al. \cite{10935624} utilized imitation learning to distill a standard cascaded PID policy into a Spiking Neural Network (SNN) running on a Cortex-M7. By training the SNN to mimic the PID output (with time-shifted targets to mitigate delay), they achieved a tracking error comparable to the classical baseline ($3.0^\circ$ vs $2.7^\circ$), effectively demonstrating that the stability of classical linear control can be preserved within the sparse, event-driven architecture of neuromorphic solvers.

To address the singularities inherent in linear methods, geometric control has been adapted for nano-UAVs \cite{greiff2017modelling}. Instead of relying on local Euler angle coordinates, this formulation defines error dynamics directly on the Special Euclidean Group $SE(3)$, specifically using the rotation matrix $R \in SO(3)$. This global definition avoids singularities entirely, allowing the controller to track aggressive trajectories (e.g., flips, loops, and inverted flight) that would saturate or destabilize a linear controller. Experimental validations confirm that while geometric controllers are more sensitive to motor noise than PIDs, they provide superior tracking accuracy for dynamic trajectories where the small-angle assumption is violated \cite{greiff2017modelling, eschmann2024learning}.

To manage the high sensitivity of nano-drones to aerodynamic disturbances (e.g., wind gusts, ground effect), Incremental Nonlinear Dynamic Inversion (INDI) has emerged as a robust alternative to standard feedback linearization \cite{van2020board, muller2023robust}. Unlike standard nonlinear dynamic inversion, which relies entirely on a model to cancel nonlinearities, INDI uses angular acceleration feedback to incrementally update the control input based on the change in the system state. By substituting instantaneous sensor measurements for model-based predictions of current aerodynamic forces, INDI becomes inherently robust to model mismatches and external disturbances. Comparative studies demonstrate that INDI achieves superior disturbance rejection compared to PID and geometric controllers, specifically in maintaining stability under turbulent conditions, although it necessitates high-quality, low-latency angular acceleration estimation to prevent noise amplification \cite{eschmann2024learning}. Additionally, L1 Adaptive Control has been successfully employed as an augmentation to baseline controllers. It separates the estimation of uncertainties from the control loop bandwidth, allowing a nano-drone to adapt to significant changes in mass or aerodynamic drag (e.g., carrying a payload) without re-tuning, a critical capability for deployment in varying environmental conditions \cite{bauersfeld2021neurobem}.

\subsubsection{Learning-Augmented Control}
While classical control laws provide stability in nominal conditions, they often fail to capture the complex aerodynamic interactions inherent to nano-UAV flight, such as rotor-to-body interference and transient blade flapping. To bridge the gap between simplified physical models and the complex reality of agile flight, the field has bifurcated into two distinct learning paradigms: hybrid residual learning and end-to-end policy synthesis.

Standard quadrotor control relies on simplified first-principles models, such as Blade Element Momentum (BEM) theory, which assumes thrust and torque are proportional to the square of the rotor speed. While adequate for hover, these models deteriorate rapidly during high-speed flight ($>18$ m/s) or aggressive maneuvers, as they neglect transient aerodynamic effects. To address these deficits without incurring the computational cost of computational fluid dynamics, NeuroBEM employs a hybrid architecture \cite{bauersfeld2021neurobem}. It does not replace the flight controller; rather, it augments the standard BEM physics engine. The total force and torque acting on the drone are modelled as the sum of nominal physics-based predictions ($f_{prop}, \tau_{prop}$) and a learned residual component ($f_{res}, \tau_{res}$).

A lightweight neural network (typically a temporal convolutional network with approximately 25K parameters) functions as a residual regressor, accepting a history of state variables (e.g., linear velocity, angular velocity, and motor speeds) to predict the aerodynamic error between the BEM model and physical reality. This architecture allows the onboard controller to plan manoeuvres using a high-fidelity dynamic model that accounts for complex aerodynamic drag and prop-wash interference. Crucially, since the hybrid model relies on BEM theory for the dominant dynamics (80--90\% of forces) and utilizes the network only to correct the remaining 10--20\% error, it exhibits superior generalization compared to ``black box'' end-to-end deep learning methods. The network fits comfortably within the 512 kB L2 memory of the GAP8 SoC and executes in approximately 100 $\mu s$, enabling the high-frequency control loops necessary for agile flight \cite{bauersfeld2021neurobem}. More importantly, the efficacy of this hybrid approach relies on a rigorous ``Sim-to-Real'' training workflow that leverages real-world flight data to bridge the reality gap. Data is collected using precise motion capture systems (e.g., Vicon) to record the ground truth state of the drone during aggressive manoeuvres \cite{ostovar2022nano}. The training targets for the neural network are generated by subtracting the forces/torques predicted by the nominal BEM model from the ground truth forces derived from the motion capture data. The network is then trained via supervised learning to minimize the prediction error of these residuals. Once trained, the network is frozen and deployed onboard to provide real-time aerodynamic corrections to the flight controller.

\subsubsection{Neuromorphic Control}
A radical shift in low-level control is the application of neuromorphic computing and spiking neural networks (SNNs) to flight stabilization. Traditional artificial neural networks struggle to efficiently replicate the integral and derivative terms of PID controllers due to the discrete, time-stepped nature of standard inference. To bridge this gap, researchers have developed input-weighted threshold adaptation for leaky integrate-and-fire neurons \cite{stroobants2023neuromorphic}. In this architecture, synaptic connections do not merely inject current; they also adapt the firing threshold of the post-synaptic neuron. This mechanism allows the SNN to effectively ``store'' an accumulated error state (integration) without explicit memory buffers, enabling zero steady-state error control akin to the Integral term in a PID controller.

The primary advantage of this paradigm is the potential for ultra-low latency and energy efficiency. SNN controllers can operate in an event-driven manner, processing sensor data only when changes occur (sparse processing). This allows them to operate at frequencies far exceeding standard MCUs (e.g., $>500$ Hz) while consuming milliwatts of power on specialized neuromorphic hardware like Intel's Loihi or in mixed-signal implementations \cite{stroobants2023neuromorphic}. However, recent research challenges the strict necessity for specialized silicon. Stroobants et al. \cite{10935624} demonstrated the first end-to-end SNN for attitude estimation and control running entirely on a standard Cortex-M7 microcontroller (Teensy 4.0). By employing a modular architecture trained via imitation learning—and crucially, using time-shifted targets to compensate for implicit synaptic delays—they achieved stable 500 Hz flight on the Crazyflie platform. This work highlights that the computational advantage of SNNs lies in replacing expensive floating-point multiplications with sparse integer additions, a benefit that translates effectively to general-purpose embedded hardware even prior to the adoption of neuromorphic chips. While validated in flight tests, neuromorphic control currently remains an emerging technology rather than a standard tool. The requirement for specialized hardware or emulation often outweighs the efficiency gains when deployed on standard off-the-shelf nano-drones compared to highly optimized classical PIDs.

\subsection{Nano-UAVs Swarm Coordination}
The coordination of nano-UAV fleets represents a paradigm shift from single-agent autonomy to collective intelligence, enabling complex missions such as search-and-rescue and heavy payload transport that exceed the capabilities of individual sub-50g platforms. However, the transition from singular to collective operation imposes severe penalties on the limited SWaP budget of nano-drones, specifically regarding communication power consumption and the onboard computational overhead required for conflict resolution.

\subsubsection{Centralized and Decentralized Architectures}
The literature presents a dichotomy between centralized architectures, which prioritize precision and trajectory optimality, and decentralized architectures, which prioritize robustness and scalability. The centralized paradigm is exemplified by the Crazyswarm architecture, which leverages the Robot Operating System (ROS) to successfully coordinate up to 49 nano-quadcopters in dense formations \cite{7989376}. This approach offloads the heavy computational burden of trajectory planning and collision avoidance to an external base station (PC), broadcasting setpoints to agents via radio. While this minimizes the onboard computational load (allowing the STM32 MCU to focus purely on attitude control) it introduces a critical Single Point of Failure (SPOF) and relies heavily on external infrastructure, such as VICON motion capture systems, for state estimation. Furthermore, this architecture faces severe bandwidth bottlenecks; telemetry latency grows linearly with swarm size, reaching approximately 26 ms for 49 vehicles due to radio bandwidth saturation, effectively limiting deployment to controlled, instrumented environments.

\begin{figure*}[t!]
    \centering
    \includegraphics[width=0.9\textwidth]{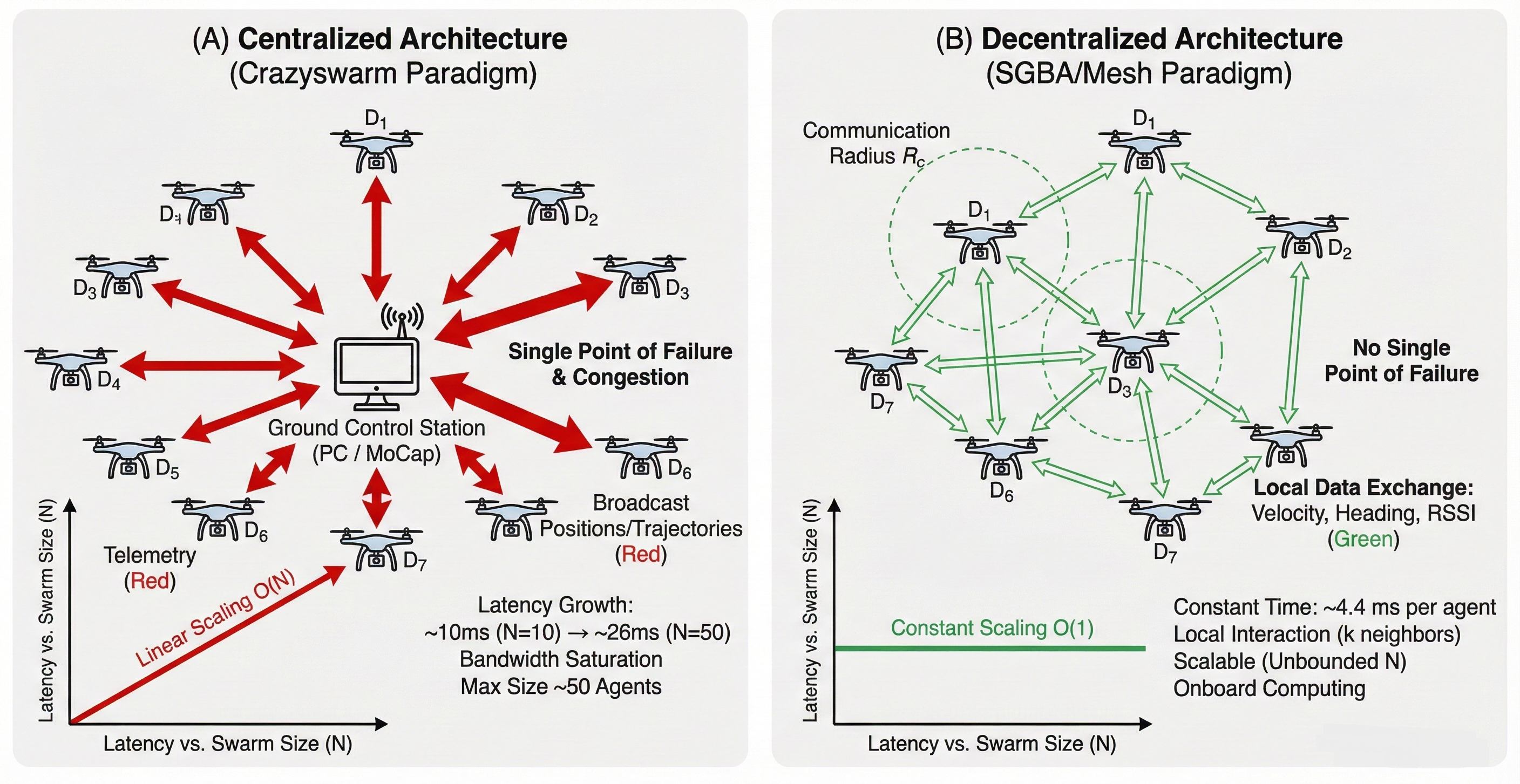}
    \caption{\textbf{Swarm Communication Architectures.} (A) Centralized paradigms (e.g., Crazyswarm) rely on a single ground station, leading to linear latency growth ($O(N)$) and single points of failure. (B) Decentralized mesh paradigms (e.g., swarm gradient bug algorithm) utilize local communication, achieving constant time complexity ($O(1)$) and enabling scalable, robust swarms in GPS-denied environments.}
    \label{fig:swarm_arch}
\end{figure*}

This scalability bottleneck is visualized in Fig. \ref{fig:swarm_arch}, which contrasts the linear latency scaling of centralized systems (Panel A) against the constant-time complexity of decentralized mesh architectures (Panel B). By utilizing local communication, decentralized approaches avoid the radio bandwidth saturation that cripples large-scale centralized swarms.
Conversely, decentralized approaches distribute the computational load across the swarm. Algorithms such as the swarm gradient bug algorithm enable swarms to explore unknown environments without external infrastructure by relying on onboard sensing and local communication for collision avoidance \cite{mcguire2019minimal}. While decentralization mitigates the SPOF problem and communication latency bottlenecks associated with central servers \cite{qazavi2023distributed}, it imposes significant challenges on the agent's SWaP constraints. Each agent must perform state estimation, mapping, and neighbour avoidance onboard, often saturating the limited RAM (e.g., 192 kB on the Crazyflie) and processing power of the microcontroller \cite{qazavi2023distributed}. To bridge this gap, frameworks like HDSM run path planning and trajectory generation on separate threads to handle communication delays, achieving flight speeds 67\% faster than comparable decentralized methods \cite{toumieh2024high}.

\subsubsection{The Cost of Coordination: Bandwidth and Energy}
The implementation of collective behaviours incurs significant costs regarding communication bandwidth and energy, which are critical for nano-UAVs. Scaling swarms reveals severe bandwidth bottlenecks. In centralized systems like Crazyswarm, telemetry latency grows linearly with swarm size, requiring specialized compressed broadcast protocols to manage 49 agents \cite{7989376}. In decentralized systems performing distributed mapping, the bandwidth required for sharing pose graphs also scales linearly; for example, a swarm of 20 drones utilizing P2P communication consumes approximately 64 kbit/s, saturating standard radios \cite{friess2024fully}.

The feasibility of swarm coordination on nano-UAVs is dictated by the trade-off between algorithmic sophistication and hardware limitations. While centralized methods remain the high-performance feasible standard for precision in controlled environments, they are undeployable in the wild. Decentralized optimization (e.g., AMSwarm) and Bio-inspired heuristics are edge-feasible, provided that communication frequency is minimized to conserve battery. However, reliable collaborative transportation and large-scale dense swarming remain emerging / edge-restricted, constrained primarily by the high-power consumption of ranging sensors (UWB) and the exponential complexity of resolving inter-agent collisions in tight spaces without a central clock. 

\subsection{Software Ecosystem and Toolchains}
The development pipeline for nano-UAVs necessitates a rigorous transition from high-level algorithmic design to efficient binary execution on severely constrained hardware. This section analyses the critical software stack enabling this transition, evaluating both simulation environments that bridge the ``Sim-to-Real'' gap and the compilation toolchains and middleware that address the ``Memory Wall,'' with a consistent focus on computational latency, memory footprint, and physical fidelity.

\subsubsection{Simulation Environments: Bridging the Reality Gap}
Given the physical fragility of nano-UAVs and the prohibitive cost of real-world data collection, high-fidelity simulation has become a mandatory prerequisite for deployment. The ecosystem is currently bifurcated into two distinct classes: simulators that prioritize photorealistic rendering to minimize the ``visual reality gap'' for perception networks, and those that prioritize computational throughput to accelerate the convergence of RL policies.

For tasks relying on computer vision, such as monocular depth estimation or semantic segmentation, minimizing the domain shift between synthetic training data and real-world imagery is critical \cite{mairaj2019application}. Microsoft AirSim serves as the standard in this domain, leveraging the Unreal Engine to provide photorealistic rendering and precise physics. This fidelity makes it the preferred tool for training AI backbones where lighting, texture, and shadow artifacts must be rigorously modelled to prevent overfitting. However, the computational load of AirSim is substantial, often requiring high-end GPUs to function effectively \cite{panerati2021learning}. This heavy footprint creates a bottleneck for sample-inefficient RL algorithms, limiting the ability to run the massive parallel simulations required for policy convergence.

In scenarios involving multi-agent coordination, the simulation challenge shifts from visual fidelity to the efficient management of massive state spaces. ARGoS \cite{stolfi2024argos} and Webots \cite{mcguire2019minimal} have emerged as the industry standards for these ``swarm-centric'' workflows. ARGoS features a modular, multi-threaded architecture capable of simulating large fleets efficiently, with recent plug-ins successfully modeling the specific electro-mechanical dynamics of the Crazyflie 2.1, including battery discharge curves and AI-deck sensor noise. Similarly, Webots enables the procedural generation of randomized environments, a feature heavily utilized in developing exploration logic like the SGBA algorithm to validate coverage metrics across diverse topologies before physical deployment.

To address the ``Sim-to-Real'' gap in flight control, simulation must capture the specific aerodynamic non-linearities of nano-scale flight while maintaining high throughput. Gym-pybullet-drones \cite{panerati2021learning} has emerged as a critical tool for this niche. Built on the Bullet Physics engine and integrating an OpenAI Gym interface, it explicitly models aerodynamic interaction effects often ignored by general solvers—such as ground effect and downwash interactions between proximate drones—while achieving update rates of 240 Hz with speed-ups exceeding 15 times wall-clock time on standard laptops. This efficiency facilitates the generation of the millions of samples required for algorithms like PPO and DDPG. While other engines like MuJoCo excel in contact-rich dynamics \cite{stolfi2024argos}, they typically require extensive parameter tuning to accurately represent rotorcraft aerodynamics \cite{akbari2024tiny}. Furthermore, specialized simulators like SUAAVE focus on the interdependence of UAV flight and ad-hoc networking, modelling the communication constraints inherent in distributed swarms \cite{mairaj2019application}. Similarly, FlightGear offers robust flight dynamics modeling suitable for R\&D, but lacks the specific sensor models (e.g., low-resolution ToF matrices) and swarm interfaces required for modern nano-UAV research \cite{mairaj2019application}.

A comparative feasibility analysis of these simulation and middleware tools highlights their distinct roles within the development pipeline. For RL control policies, Gym-pybullet-drones establishes the industry standard by optimally balancing computational speed for training with the physical fidelity of aerodynamics critical for Sim-to-Real transfer \cite{mairaj2019application}. In contrast, for validating decentralized swarm behaviors, platforms like ARGoS and Webots are essential due to their high-performance simulation capabilities that scale to large numbers of agents \cite{mcguire2019minimal, stolfi2024argos}. While Microsoft AirSim provides superior visual realism for perception-based research, its high computational overhead restricts its feasibility for training large-scale swarm policies or constrained hardware-in-the-loop setups \cite{panerati2021learning}. Finally, although middleware frameworks such as ROS/Python are excellent for prototyping, their standard interfaces introduce latencies unacceptable for high-frequency onboard control, necessitating eventual compilation to lower-level languages like C/C++ or Rust for deployment \cite{bohmer2020latency}.

\subsubsection{Middleware and Interfaces: The Latency-Footprint Trade-off}
The software interface between high-level control logic and the low-level flight controller is a critical determinant of system responsiveness and energetic efficiency. Within standard research platforms like the Bitcraze Crazyflie ecosystem, rapid prototyping is typically conducted using the Python-based client library (cflib), which facilitates easy integration with machine learning frameworks such as PyTorch \cite{10611492}. However, this convenience comes at a significant performance cost. Bohmer et al. \cite{bohmer2020latency} demonstrated that the interpreter and serial communication overhead introduce non-deterministic latency, measuring round-trip times (RTT) of approximately 18 ms. By translating these critical path components to system-level languages like Rust, they reduced this latency to $\sim$9 ms, approaching the theoretical minimum imposed by the radio hardware and underscoring the necessity of compiled binaries for deployment.

For multi-agent systems, the Crazyswarm package provides a standardized, centralized architecture for swarm coordination, managing up to 49 nano-UAVs via ROS \cite{7989376}. While effective for synchronized flight demonstrations, this paradigm depends on external infrastructure (e.g., VICON motion capture) for state estimation, broadcasting pose data to the entire swarm from a single base station. This creates an inherent bottleneck; latency scales linearly with swarm size due to radio bandwidth saturation, reaching about 26 ms for 49 agents, establishing a hard scalability limit and a single point of failure. Consequently, the field is witnessing a distinct shift toward decentralized architectures that push computation to the edge. Supporting this trend, middleware such as the Crazybridge enables the integration of lightweight companion computers (e.g., Raspberry Pi Zero W), though maintaining low-latency, reliable communication for control still requires careful protocol selection, such as employing Predictably Reliable Real-time Transport (PRRT) over standard TCP/IP stacks \cite{bohmer2020latency}.

\subsubsection{Deployment and Compilation Toolchains}
The transition of DNNs from training frameworks to onboard execution is bottlenecked by the ``Deep Learning Memory Wall''. While training exploits gigabytes of memory and floating-point precision on workstations, inference on nano-UAVs is confined to microcontrollers with limited SRAM (typically $<$1 MB) and power budgets below 100 mW \cite{9381618}. Bridging this disparity requires specialized deployment compilers that automate quantization, memory tiling, and operator fusion, distinct from general-purpose libraries like TensorFlow Lite.

For standard ARM Cortex-M microcontrollers (e.g., STM32F4/ESP32), the deployment ecosystem is split between generic interpreters and vendor-specific compilers. TensorFlow Lite for Microcontrollers (TFLM) serves as the standard entry point \cite{sartori2023autonomous}, facilitating model compression via quantization (converting 32-bit floats to 8-bit integers) to reduce memory footprints by a factor of 4 \cite{kaur2023survey, palossi2021fully}. However, TFLM’s reliance on an interpreter-based execution model precludes aggressive optimizations like layer fusion, limiting performance. To mitigate this overhead on specific hardware like the Crazyflie's STM32 chips, the X-CUBE-AI stack converts models directly into optimized C libraries that utilize hardware caches rather than runtime parsing \cite{9381618}. Yet, even with this compilation, standard MCU architectures trail behind specialized PULP systems; benchmarks indicate that DORY on a GAP8 achieves 12.6 times higher energy efficiency and 7.1 times higher performance than X-CUBE-AI on a comparable STM32H7, highlighting the inherent efficiency gap between cache-based and scratchpad-based memory management \cite{9381618}.

The GreenWaves GAP8/9 SoCs, capable of Parallel Ultra-Low Power (PULP) processing, eschew hardware-managed caches in favor of explicitly managed scratchpad memories (L1 and L2). This architecture demands software that manually orchestrates data movement, a task handled by dedicated code generators. The vendor-supplied solution, GreenWaves' proprietary \textit{AutoTiler}, analyzes neural network graphs to generate C code that fragments tensors into ``tiles'' fitting within the 64 kB L1 memory. Crucially, it implements software pipelining: while cluster cores compute operations on the current tile, the Direct Memory Access (DMA) engine asynchronously fetches the next, effectively hiding access latency \cite{niculescu2021improving}. Extending this paradigm to open-source workflows, \textit{DORY} (Deployment ORiented to memorY) reformulates memory management as a constraint programming  problem \cite{9381618}. Unlike the standard AutoTiler, DORY solves for optimal tile sizes to maximize L1 utilization and utilizes the PULP-NN backend—a library of kernels optimized for RISC-V SIMD instructions. Comparative studies demonstrate that this rigorous optimization allows DORY to achieve up to 2.5 times higher throughput than vendor-supplied tools on pointwise convolutions by better exploiting data layout \cite{9381618}.

Addressing the unique requirements of RL, BackpropTools is a header-only C++ library designed for high-performance training and inference on embedded devices \cite{eschmann2023backproptools}. Unlike standard deep learning compilers that focus solely on inference, BackpropTools avoids dynamic memory allocation at runtime and integrates tightly with simulation environments. This enables ``TinyRL''—the execution of backpropagation directly on microcontrollers like the Teensy 4.0  \cite{eschmann2023backproptools}—reducing the iteration time for learning simple stabilization tasks to mere seconds \cite{eschmann2024learning}.

In terms of deployment feasibility, the ecosystem is stratified by the trade-off between performance, openness, and utility. GAPflow (AutoTiler) serves as the industry standard, offering a robust and vendor-supported workflow that ensures reliability, though its partially closed-source nature limits deep customization. In contrast, DORY paired with PULP-NN is classified as high-performance feasible; it delivers superior throughput and research flexibility for maximizing hardware utilization, albeit at the cost of complex configuration \cite{9381618}. For general-purpose microcontrollers, TFLM provides a standard feasible baseline accessible to broad hardware, yet it suffers from performance bottlenecks due to interpreter overhead \cite{sartori2023autonomous, kaur2023survey}. Consequently, X-CUBE-AI remains the preferred high-performance feasible solution for standard STM32 flight controllers, bridging the gap by compiling networks directly to optimized C code \cite{9381618}. Finally, BackpropTools represents an Emerging paradigm, distinguished by its unique capability for on-device training which is essential for realizing adaptive control policies in the field \cite{eschmann2023backproptools}.

\section{Key Challenges and Constraints}
\label{sec:challenges}

\begin{figure*}[t!]
    \centering
    \includegraphics[width=0.95\textwidth]{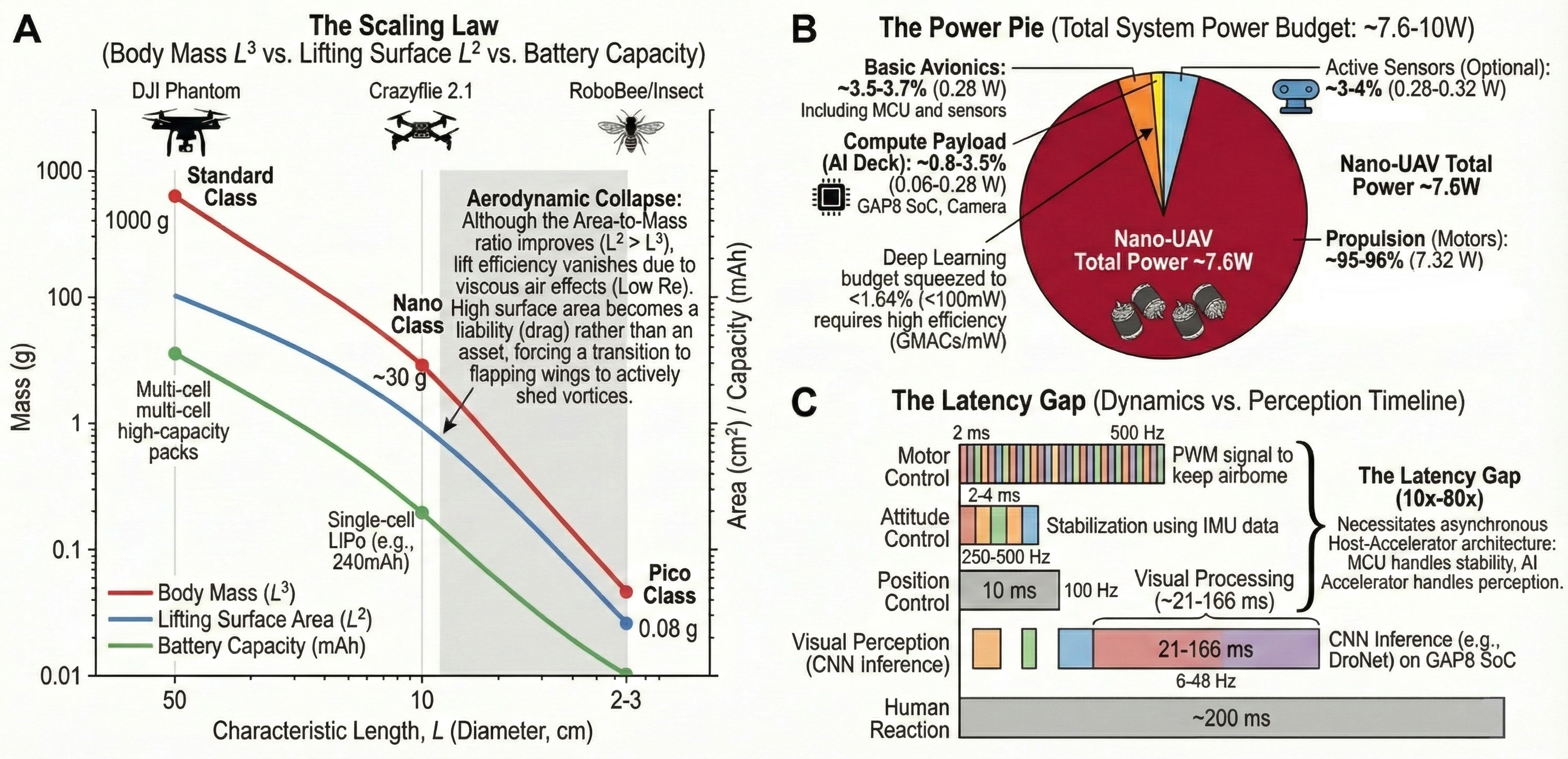}
    \caption{\textbf{System-Level Constraints Governing Nano-UAV Autonomy.} 
    \textbf{(A) Physics of scaling:} As the characteristic length $L$ decreases, mass scales with $L^3$ while lifting surface area scales with $L^2$, forcing operation in progressively lower aerodynamic efficiency regimes. 
    \textbf{(B) Power budget allocation:} Propulsion dominates the energy budget ($>$95\%), leaving less than 100\,mW for onboard sensing, computation, and control (the ``5\% rule''). 
    \textbf{(C) Latency mismatch:} The fast mechanical dynamics of nano-rotors require control rates on the order of 500\,Hz, which is several orders of magnitude faster than the inference latency of vision-based CNNs, motivating hierarchical control architectures.}
    \label{fig:design_fig}
\end{figure*}

The transition of autonomous navigation from standard-sized quadrotors to nano-UAVs is not a mere scale reduction; it constitutes a fundamental redefinition of the operational envelope governed by cubic scaling laws. These scaling relations give rise to tight, interdependent trade-offs among flight endurance, sensing capability, and onboard computational resources. In this section we quantify those constraints and show how they motivate the specialised hardware architectures described in Section~\ref{sec:hardware} and the aggressive algorithmic compression techniques presented in Section~\ref{sec:algorithms}. We encapsulate these systemic bottlenecks in Figure~\ref{fig:design_fig}, which illustrates the ``scaling paradox'': a domain where geometric advantages are negated by viscous aerodynamics and energy density limits, forcing a hard compromise between flight time and intelligence.

\subsection{The Physics of Scaling: The Energy-Autonomy Trade-off}
While the Introduction defined the ``Physics Gap'' governing nano-scale flight, its practical consequence is a severe energy-autonomy trade-off. As the characteristic length of the drone decreases to the sub-10 cm range \cite{valente2024heterogeneous}, the cubic scaling of mass versus the quadratic scaling of lifting surface area results in a precipitous drop in battery capacity (see Figure~\ref{fig:design_fig}A). For a typical 27--33 g platform like the Crazyflie 2.x, the total power budget is approximately 7--10 W \cite{palossi2021fully, e2fbb835-8278-3148-8a69-8aad65c23c4e}. However, aerodynamic lift consumes the vast majority of this budget; motors typically draw 95--96\% of the total energy to maintain flight \cite{palossi2021fully}. This physical reality leaves a minuscule power envelope for the avionics and onboard intelligence, typically bounded between 5\% and 15\% of the total system power \cite{valente2024heterogeneous, niculescu2021improving}. Quantitatively, this results in a strict compute budget of sub-100 mW to maximum 200--300 mW for all sensing, processing, and control tasks\cite{valente2024heterogeneous, niculescu2021improving}. This constraint creates a direct causal chain affecting autonomy: Every milliwatt consumed by the navigation engine subtracts directly from the flight time. For example, adding a dedicated visual navigation engine like the PULP-Shield (running the DroNet CNN) reduces the flight time by approximately 22\% (from $\sim$440 s to $\sim$340 s) due to the combined effect of added payload weight (5 g) and electrical consumption \cite{palossi201964, crupi2024fusing}. Consequently, the high-performance GPUs used in micro-aerial vehicles (e.g., NVIDIA Jetson, consuming $>$10W) are physically impossible to deploy on nano-drones, necessitating the ultra-low-power microcontroller-based strategies discussed in Section ~\ref{sec:hardware}.

\subsection{The Deep Learning Memory Wall}
The most severe bottleneck preventing the deployment of off-the-shelf deep learning models is the extreme scarcity of onboard memory. While state-of-the-art CNNs like ResNet-101 or EfficientNet require hundreds of megabytes for weights and activations, nano-UAV microcontrollers operate in a kilo-byte regime \cite{9381618}. Standard MCUs (e.g., STM32F4) feature roughly 192 kB of SRAM \cite{valente2024heterogeneous, e2fbb835-8278-3148-8a69-8aad65c23c4e}. Even advanced parallel ultra-low-power (PULP) SoCs like the GAP8 are constrained to 512 kB of L2 memory and 64 kB of L1 scratchpad memory \cite{9381618}.This creates a ``Memory Wall'' where moving data becomes more energy-expensive than the computation itself. For instance, a standard MobileNetV1 architecture, often considered ``lightweight'', requires memory bandwidths that exceed the capabilities of low-power interfaces if not aggressively optimized. This constraint forces specific algorithmic compromises: the impossibility of fitting full-precision (float32) weights into 512 kB L2 memory necessitates the 8-bit quantization and batch-normalization folding techniques used in tools like DORY. Furthermore, the lack of hardware caches on these specialized SoCs requires explicit, software-managed DMA transfers (tiling) to move data between off-chip L3 RAM (e.g., HyperRAM) and on-chip L1 memory without stalling the processing cores.

\subsection{The Latency Gap: Computational Throughput vs. Flight Dynamics} 
The operational reality of nano-UAVs is governed by a critical disconnect between the mechanical stability requirements of the platform and the computational throughput of onboard perception engines. Due to their low mass ($<$33g) and small rotational inertia ($I_{xx} \approx 1.4 \times 10^{-5} \text{ kg m}^2$) \cite{luis2016design}, nano-quadrotors exhibit extremely fast, unstable dynamics that require high-bandwidth feedback control. The standard attitude control loop on the Crazyflie 2.1 must operate at 500 Hz to maintain stability \cite{7989376, huang2023dattdeepadaptivetrajectory, greiff2017modelling}, while the state estimator typically updates at 1 kHz \cite{lambert2019low}. However, generating these control signals via visual perception on MCU-class processors is bounded by limited instruction throughput. A basic obstacle avoidance CNN requires $\sim$41 million multiply-accumulate (MAC) operations per frame \cite{palossi2021fully}, which would result in unacceptably high latency on a standard 168 MHz Cortex-M4. This throughput deficit justifies the use of heterogeneous accelerators (e.g., GAP8), which can achieve peak throughputs of $\sim$10 GMAC/s/W \cite{flamand2018gap}. Yet, even with acceleration, complex CNNs like DroNet operate at only 6--18 frames per second (FPS) \cite{palossi201964, cereda2021improving}, yielding a perceptual latency between 55 ms and 160 ms.

This creates a dangerous ``latency gap'' (see Figure~\ref{fig:design_fig}C): the drone must stabilize itself 30 to 80 times for every single visual update it receives. The causal chain of instability (Low Inertia $\to$ High-Frequency Dynamics ($>$500Hz) $\to$ Perceptual Latency ($>$50ms)) necessitates the hierarchical control architectures. High-level visual navigation engines cannot directly drive the motors; instead, they must generate setpoints (e.g., velocity or attitude commands) for a lower-level, high-frequency PID controller running on the MCU \cite{palossi2021fully, lambert2019low}. Consequently, any interruption in the perception loop or communication link exceeding 20--40 ms can lead to catastrophic instability \cite{van2020board,bohmer2020latency}. However, emerging neuromorphic architectures are beginning to close this gap. Stroobants et al. \cite{10935624} demonstrated that by replacing standard CNNs with SNNs for attitude control, the perception-actuation loop can operate at 500 Hz on a standard microcontroller, matching the frequency of the inner control loop and effectively eliminating the latency penalty associated with deep learning inference.

\subsection{Aerodynamic Instability and Sensory Noise}
The instability of nano-UAVs is further exacerbated by their susceptibility to aerodynamic disturbances that larger vehicles passively reject. Experimental data indicates that ground effect significantly increases thrust when hovering at low altitudes, requiring specific compensation coefficients in the control loop to prevent vertical instability \cite{panerati2021learning}. Furthermore, in swarm configurations, the downwash from a peer drone causes a reduction in lift for agents below, necessitating safety radii of up to 0.6 m to avoid destabilization \cite{lucahigh}. Simultaneously, the high rotational speed of micro-motors (up to 2500 rad/s \cite{greiff2017modelling}) introduces significant high-frequency vibration, which propagates through the rigid frame to the sensors. Fast fourier transform analysis of load cell data reveals noise concentration specifically at the rotational frequencies of the rotors \cite{forster2015system}. This vibration severely degrades the signal-to-noise ratio of onboard MEMS IMUs and accelerometers. For instance, flapping-wing nano-robots experience ``colored noise'' at integer multiples of the flapping frequency (e.g., 21.8 Hz), requiring specialized bandstop filters to recover usable state estimates \cite{park2023development}. Similarly, ultrasonic sensors used for obstacle avoidance are susceptible to motor noise; experimental characterization shows that standard deviation in readings can decrease by 50\% only after applying exponential moving average filtering, introducing a further latency trade-off \cite{muller2024batdeck}.

\subsection{Environmental Constraints: Navigation in Cluttered Spaces}
The ability to navigate cluttered environments is strictly bounded by the limited range and FoV of lightweight sensors. Unlike larger drones equipped with 360-degree LiDAR or depth cameras, nano-UAVs often rely on sparse ToF sensors (e.g., VL53L5CX) with a maximum effective range of 4 meters and a narrow FoV of 45--65 degrees \cite{friess2024fully, muller2023robust}. This sensory myopia creates critical vulnerabilities: dynamic obstacles can enter the drone's trajectory from outside the narrow FoV faster than the perception-actuation loop (latency 210 $\mu$s for processing \cite{muller2023robust}) can react. Furthermore, environmental material properties pose a lethal threat; multi-zone ToF sensors fail to detect transparent surfaces (glass) or highly absorptive materials (black matte surfaces) at oblique angles, leading to collisions \cite{muller2024batdeck}. In symmetric environments like corridors, sparse sensing leads to aliasing, where the drone cannot distinguish between different locations, making reactive planners susceptible to local minima \cite{ostovar2022nano,zhou2022efficient}.

\subsection{Bandwidth and Communication: The ``Off-Board Myth''}
A prevalent assumption in early robotics research was that computationally intensive tasks could simply be offloaded to powerful ground stations. For nano-UAVs, the literature proves this to be an ``Off-Board Myth,'' rendered infeasible by the physics of radio communication and the overhead of standard protocols. The standard Crazyradio PA link (2.4 GHz) is throttled by the Crazy Real-Time Protocol (CRTP), which restricts payloads to 31 bytes per packet \cite{de2011modeling, sartori2023autonomous}. This fragmentation imposes a severe bottleneck on swarm scalability; in centralized architectures, telemetry latency grows linearly with swarm size. For a swarm of 49 drones, the system requires three separate radios to maintain a latency below 26 ms \cite{7989376}. In decentralized peer-to-peer (P2P) scenarios, a swarm of 20 drones sharing pose graphs consumes approximately 64 kbit/s, saturating the P2P broadcast capacity \cite{friess2024fully}. Crucially, streaming raw video for off-board processing is impossible. Streaming images via WiFi introduces variable round-trip latencies ranging from 120 ms to over 320 ms \cite{sartori2023autonomous}, orders of magnitude too slow for 500 Hz control loops. Furthermore, high-frequency radio transmission is energetically prohibitive, consuming hundreds of milliwatts \cite{palossi2021fully}, whereas onboard computation on the GAP8 SoC consumes only 64--100 mW \cite{palossi201964, cereda2021improving}. Thus, the ``Design Straitjacket'' dictates that autonomy must be moved to the edge.

\subsection{The Sim-to-Real Gap}
The disparity between simulated training environments and physical reality, known as the ``Sim-to-Real'' gap, is a primary cause of failure for learning-based controllers. This gap manifests in two distinct domains: dynamics mismatch and visual domain shift. Simulators often model quadrotors as rigid bodies with immediate actuator response \cite{sacks2022learning}. However, physical nano-drones exhibit significant actuator delays; experimental system identification reveals a motor time constant of approximately $\tau \approx 0.15$ s, a delay significantly larger than the control interval of low-level loops \cite{eschmann2024learning}. Novel training strategies are emerging to explicitly compensate for this temporal mismatch. For instance, Stroobants et al. \cite{10935624} introduced a time-shifting technique where the onboard network is trained to predict the expert's output $k$ steps into the future. This effectively forces the neural network to learn an internal forward model that cancels out the inherent synaptic and actuator delays, significantly reducing tracking error and oscillations in real-world flight. When policies trained without this latency are deployed, the mismatch leads to catastrophic oscillations \cite{eschmann2024learning}. For vision-based navigation, policies trained on synthetic images fail due to differences in lighting and texture. Standard background augmentation is often insufficient; however, techniques like generalization through simulation and background randomization have been shown to reduce the mean squared error of pose estimation by nearly 40\% in unseen environments, proving that aggressive domain randomization is strictly necessary to bridge the gap between clean synthetic data and noisy, motion-blurred real-world imagery \cite{cereda2021improving, kang2019generalization, stolfi2024argos}.

\section{Applications}
\label{sec:applications}

The deployment of sub-50g nano-UAVs represents a shift from general-purpose aerial utility to specialised platforms tailored for environments where conventional micro aerial vehicles are physically excluded or operationally unsafe. In this section we analyse how the constrained form factor and the coupled SWaP and performance limitations of nano-drones delimit their utility in confined, GPS-denied, and human-populated settings, before detailing the fundamental research gaps that must be bridged to enable full autonomy.

\subsection{Operations in Restricted and Sensitive Environments}
Nano-UAVs are increasingly targeted for environments characterized by severe spatial constraints and the absence of Global Navigation Satellite System (GNSS) signals. Unlike standard quadrotors ($>$50 g) which rely on GNSS and heavy payload sensors, nano-scale platforms typically utilize their sub-10 cm form factor to traverse ``human-inaccessible'' zones such as collapsed infrastructure, subterranean mines, or the interior of complex machinery \cite{friess2024fully, uppaluru2023multi}. In search and rescue scenarios, such as post-earthquake assessment, the low inertia of these platforms (approx. 27--40 g) renders them intrinsically safe for operation in close proximity to victims or unstable debris, minimizing the risk of secondary collapse or injury \cite{muller2023robust, giernacki2017crazyflie}. To operate in these GPS-denied zones, the navigation stack must shift from absolute global positioning to relative local state estimation. Due to strict payload limits ($<$15 g), platforms utilize lightweight multizone ToF sensors (e.g., VL53L5CX) to provide sparse $8 \times 8$ pixel depth maps for obstacle avoidance \cite{niculescu2022towards,niculescu2023nanoslam}, augmented by monocular optical flow to mitigate velocity drift \cite{muller2023robust}. Advancements like NanoSLAM leverage graph-based optimization on edge processors (e.g., GAP9) to achieve 4.5 cm mapping accuracy using only onboard sensors, enabling navigation in dynamic industrial environments without external computation \cite{niculescu2023nanoslam}. However, there is a distinct trade-off regarding inspection fidelity; while nano-drones grant access to confined spaces (e.g., jet engine interiors), their low-resolution sensors (QVGA cameras or sparse ToF) often fail to resolve micro-fractures or subtle surface defects visible to larger micro aerial vehicles \cite{de2018inverted, niculescu2023nanoslam}, positioning them primarily as ``scouts'' for rapid initial assessment rather than detailed forensic analysis \cite{friess2024fully}.

This low-inertia and small form factor also enable a shift from high-altitude observation to close-range, stealthy monitoring in the surveillance and agricultural domains. Their primary operational advantage is low detectability; research indicates that nano-UAVs produce significantly lower acoustic noise ($\sim$40 dB) compared to standard quadcopters (up to 75 dB), allowing for covert operations in quiet environments \cite{crupi2025efficient}. To leverage this proximity, recent frameworks integrate lightweight deep learning models directly onboard. Systems utilizing YOLOv7 combined with long short-term memory networks have been demonstrated to detect suspicious activities (e.g., loitering) with high accuracy \cite{assudani2025autonomous}. Given the high power cost of video streaming, edge computing approaches that process video onboard and transmit only metadata are essential, reducing latency to approximately 45 ms and power consumption to 2.1 W compared to cloud-based solutions \cite{assudani2025autonomous}. Similarly, in precision agriculture, these stealth and size advantages allow nano-UAVs to serve as scouts for fine-grained pest detection, operating within crop canopies where satellite imagery lacks resolution and larger drones risk damaging plants via downwash. The integration of ultra-low-power accelerators, such as the GreenWaves GAP9, enables the execution of models like SSDLite-MobileNetV3 directly onboard. Research by Crupi et al. \cite{crupi2025efficient} demonstrates that such models can identify specific pests (e.g., \textit{Popillia japonica}) with a mean average precision of 0.79 while running at 6.8 fps. This capability facilitates heterogeneous systems where nano-drones identify infestation ``hotspots'' to optimize the path of ground-based treatment robots, potentially reducing total working time by up to 20 hours for a $200 \times 200~\text{m}^2$ field.

\subsection{Logistics and Swarm Coordination}
The application of nano-UAVs in logistics bifurcates into inventory management and physical delivery, each addressing distinct constraints through single-agent or swarm behaviors. For inventory tracking, nano-drones utilize wall-following algorithms (derived from the Swarm Gradient Bug Algorithm) to autonomously scan barcodes in narrow aisles without external positioning infrastructure \cite{arvidsson2023drone}. This autonomous scanning allows for the rapid generation of item location maps in warehouse spaces that are typically inaccessible to ground robots due to verticality or clutter. Conversely, while single nano-drones lack the payload capacity for delivery ($<$15 g), collaborative transportation offers a solution to the under-capacitated vehicle routing problem \cite{sreedhara2024optimal}. Swarms connected via cable-suspended mechanisms can lift heavier payloads by utilizing distributed trajectory optimization to reconfigure formations and pass through narrow apertures \cite{jackson2022accelerating}. However, the high energy cost of hovering with a payload severely restricts the operational radius, making this currently a theoretical optimization challenge rather than a deployable logistic solution.

Regardless of the specific logistic task, operating effectively in large environments often requires swarm collaboration to overcome field-of-view and battery limitations. Frameworks like NanoSLAM utilize graph-based SLAM running onboard the GAP9 SoC to optimize pose graphs where nodes represent drone states \cite{niculescu2023nanoslam}. In distributed scenarios, agents exchange pose graphs and align sparse scans using iterative closest point algorithms when trajectories intersect \cite{friess2024fully}. This approach scales linearly; a swarm of 4 drones can map a maze in under 3 minutes, a task that would deplete the battery of a single agent. However, memory constraints (e.g., 512 kB RAM) impose a hard limit, restricting the maximum mappable area to approximately $80 \text{ m}^2$ before the graph saturates available resources.

\subsection{High-Dynamics and Human Interaction}
The physical safety profile of nano-UAVs positions them as the primary platform for proximate Human-Drone Interaction (HDI). Unlike larger drones, their negligible mass renders them intrinsically safe for ``follow-me'' applications without safety cages. The primary challenge is Visual human pose estimation under SWaP constraints. The PULP-Frontnet CNN addresses this by regressing relative 3D pose from low-resolution ($160 \times 96$) images, achieving inference rates of up to 135 Hz on the GAP8 SoC while consuming only 87 mW \cite{palossi2021fully}. Emerging paradigms like TagTeam further reduce computational load by synchronizing drone movement with wearable sensors via dead-reckoning \cite{jayarajah2022tagteam}. While HDI leverages low inertia for safety, autonomous drone racing leverages it to test control algorithms at the absolute limits of handling. The agility of nano-drones allows them to test aerodynamic models at speeds up to 18 m/s in confined arenas. At these speeds, aerodynamic drag and rotor interactions dominate; hybrid architectures like NeuroBEM \cite{bauersfeld2021neurobem} fuse blade element momentum theory with neural networks to predict residual forces, reducing tracking error by 50\% during aggressive maneuvers. While geometric controllers generally provide lower tracking error, robust controllers like INDI offer superior rejection of aerodynamic disturbances \cite{eschmann2024learning}.

In summary, the application review reveals a hierarchy of practical feasibility determined by the coupled SWaP and performance constraints of the nano form factor. \textbf{High-performance feasible} tasks (for example, drone racing and short-duration HDI) are presently deployable: the platform’s low inertia enables aggressive manoeuvres and safe interaction when combined with advanced control stacks such as NeuroBEM and PULP-Frontnet. \textbf{Edge-feasible} domains, including collaborative mapping and pest monitoring, have been functionally validated but remain essentially ``research-grade'' due to the limited range of sparse sensors ($\approx$4 m) and severe onboard memory constraints. Finally, \textbf{theoretical} applications such as last-mile delivery are the least mature; although techniques like swarm-based collaborative lift are physically possible, the prohibitive energy cost confines practical operation to only a few metres, precluding commercial viability without major advances in energy storage or propulsion efficiency.

\section{Open Challenges}
\label{sec:open challenges}
\subsection{Fundamental Research Gaps}
Despite the progress in specific application domains, the deployment of nano-UAVs is generally hindered by a series of interdependent physical and computational constraints that create a ``Physics Gap'' between required control frequency and available power. Table~\ref{tab:open_problems} provides a summary of these challenges and the corresponding architectural solutions.

The primary constraint defining nano-UAV autonomy is the rigorous application of the square-cube law. As the characteristic length of the drone decreases, its mass and volume scale cubically, while lifting surface area scales squarely. This results in a precipitous drop in battery capacity, limiting the total power budget to approximately 7--10 W for a 30 g platform \cite{palossi2021fully, e2fbb835-8278-3148-8a69-8aad65c23c4e}. With propulsion consuming $>$95\% of this budget, the avionics envelope is compressed to sub-100 mW \cite{valente2024heterogeneous}. Simultaneously, the negligible rotational inertia ($I_{xx} \approx 1.4 \times 10^{-5} \text{ kg m}^2$) renders the platform hypersensitive to aerodynamic disturbances \cite{luis2016design}. This necessitates high-frequency control ($>$500 Hz), yet the energy budget precludes the high-speed processors required to compute those controls.

A critical disconnect exists between mechanical stability requirements and perception throughput. Generating control signals via visual perception is bounded by instruction throughput; even with accelerators like the GAP8, complex CNNs operate at only 6--18 FPS, yielding a perceptual latency of $>$55 ms \cite{palossi201964}. Consequently, the drone must stabilize itself mechanically 30--80 times for every single visual update, where any interruption leads to catastrophic divergence \cite{bohmer2020latency}. Furthermore, the ``Memory Wall'' (onboard SRAM $<$1 MB) prevents the deployment of standard deep learning backbones. The energy cost of moving data between off-chip RAM and the processor often exceeds the cost of computation itself, forcing the adoption of aggressive 8-bit quantization and depth-wise separable convolutions that trade accuracy for feasibility \cite{9381618}.

Nano-UAVs also suffer from ``Sensory Myopia'' due to reliance on sparse ToF sensors with narrow FoV (45--65$^\circ$) and limited range ($<$4 m), making them vulnerable to dynamic obstacles and perceptual aliasing in symmetric environments \cite{muller2023robust, zhou2022efficient}. Moreover, off-loading computation to ground stations is an ``Off-Board Myth''; the physics of radio communication (bandwidth saturation) and standard protocol overheads introduce non-deterministic latencies ($>$100 ms for video streaming) that violate stability criteria, necessitating that autonomy remains indigenous to the edge \cite{7989376, sartori2023autonomous}.

\begin{table*}[ht]
\renewcommand{\arraystretch}{1.3} 
\caption{Summary of Open Problems in Nano-UAV Autonomy. This table maps the fundamental constraints (Section \ref{sec:open challenges}) to required architectural shifts.}
\label{tab:open_problems}
\centering
\begin{tabularx}{\textwidth}{@{} l >{\raggedright\arraybackslash}p{3.2cm} >{\raggedright\arraybackslash}X >{\raggedright\arraybackslash}p{3.2cm} @{}}
\toprule
\textbf{Challenge} & \textbf{Current State (SoA)} & \textbf{Remaining Gap} & \textbf{Potential Solution} \\
\midrule

\textbf{SWaP} & 
Modular PCBs; Avionics power $<100$ mW. & 
\textbf{The Physics Gap:} \newline Energy budget precludes processors fast enough ($>500$ Hz) to handle low-inertia dynamics. & 
\textbf{System-in-Package:} \newline Single-die integration of flight controller, AI, and radio. \\
\addlinespace

\textbf{Latency} & 
CNNs on accelerators (e.g., GAP8) at 6--18 FPS. & 
\textbf{The Control Gap:} \newline Drone requires 30--80 mechanical corrections per visual update; interruption leads to divergence. & 
\textbf{Neuromorphic Eng.:} \newline Event-based cameras and SNNs for asynchronous, $\mu$s latency. \\
\addlinespace

\textbf{Memory} & 
Aggressive 8-bit quantization; Depth-wise convs. & 
\textbf{The Efficiency Gap:} \newline Energy cost of data movement (RAM to CPU) exceeds computation cost. & 
\textbf{Bio-inspired Arch.:} \newline Computing-in-memory and sparse processing to bypass von Neumann limits. \\
\addlinespace

\textbf{Sim-to-Real} & 
Offline training; Static policies. & 
\textbf{The Adaptation Gap:} \newline Inability to compensate for battery degradation or motor damage mid-flight. & 
\textbf{On-Chip Learning:} \newline TinyML frameworks (e.g., BackpropTools) for online fine-tuning. \\
\addlinespace

\textbf{Connectivity} & 
Off-loading compute to ground station. & 
\textbf{The Reliability Gap:} \newline Bandwidth saturation and non-deterministic latency ($>100$ ms) violate stability criteria. & 
\textbf{Indigenous Autonomy:} \newline Full edge-processing without external dependence. \\
\bottomrule
\end{tabularx}
\end{table*}

\subsection{Future Architectures and Recommendations}
To bridge the fundamental performance gaps inherent to nano-UAVs, future research must pivot from component optimization to holistic architectural shifts in both silicon design and control paradigms. The current modular approach of stacking PCBs incurs unacceptable weight penalties, necessitating a convergence toward System-in-Package (SiP) solutions that integrate the flight controller, AI accelerator, and radio onto a single die to minimize interconnect weight and maximize performance-per-watt \cite{valente2024heterogeneous, muller2023fully}. Simultaneously, to close the latency gap, control loops should transition from discrete-time processing to event-based architectures; SNNs running on neuromorphic hardware can process sensor changes asynchronously with microsecond latency, matching the temporal dynamics of the platform \cite{stroobants2023neuromorphic}. Initial validation of this paradigm is already visible, with recent work successfully deploying end-to-end spiking control policies on commercial microcontrollers (Teensy 4.0) via imitation learning, proving that the benefits of event-based processing (specifically the reduction of floating-point overhead) can be realized even before the widespread availability of specialized neuromorphic hardware \cite{10935624}.  Finally, to address the ``Sim-to-Real'' gap, agents must possess the capacity for adaptation post-deployment; emerging frameworks like BackpropTools demonstrate the feasibility of running backpropagation directly on microcontrollers, allowing drones to fine-tune control policies in the field to compensate for battery degradation or motor damage \cite{eschmann2023backproptools}.

\section{Conclusion}
The development of sub-50g autonomous nano-UAVs constitutes one of the most exacting engineering challenges in contemporary robotics. As this review has shown, moving to the nano-scale is not a straightforward down-scaling of existing systems but a transition to a distinct operational regime governed by tightly coupled the SWaP and performance constraints. Under these scaling laws, every milligram of payload and every milliwatt of power must be traded directly against flight endurance and sensing capability. We synthesise an onboard autonomy stack that enables operation in this regime, identifying Parallel Ultra-Low-Power (PULP) architectures, aggressively quantised deep neural networks, and hybrid learning-based control as its principal components. The current evidence indicates that short-term visual navigation in controlled environments has been largely addressed: compressed perception pipelines (e.g., DroNet, PULP-Frontnet) can run within the sub-100\,mW envelope of edge processors, and modern flight controllers (e.g., INDI, NeuroBEM) close the loop on agile, stable flight.

Nevertheless, incremental optimisation of conventional robotics primitives is approaching practical limits. The observed latency gap between fast rotor dynamics and visual-inference times, together with the platform memory wall, imply that further gains will become increasingly marginal unless the community adopts fundamentally different computational paradigms. Promising directions include event-based sensing and neuromorphic or bio-inspired architectures, which emulate the extreme efficiency of small biological nervous systems and offer a path to bypass the inefficiencies of conventional von Neumann pipelines. If these architectural shifts materialise, they could unlock a new frontier for nano-UAVs: decoupling meaningful autonomy from bulky infrastructure and enabling pervasive, low-risk operation in confined, GPS-denied, or human-populated environments. Applications ranging from search-and-rescue to precision agriculture would thereby benefit from truly pervasive, low-cost aerial sensing platforms.

\newpage








\end{document}